\documentclass[a4paper,fleqn]{cas-dc}

\usepackage[utf8]{inputenc}
\usepackage[numbers]{natbib}
\usepackage{hyperref}
\usepackage{amsmath,amssymb,amsfonts,amsthm}%
\usepackage{bm}%
\usepackage{graphicx}%
\usepackage{cas-common}
\usepackage[normalem]{ulem}

\theoremstyle{plain}
\theoremstyle{plain}%
\theoremstyle{plain}%
\theoremstyle{plain}

\def\tsc#1{\csdef{#1}{\textsc{\lowercase{#1}}\xspace}}
\tsc{WGM}
\tsc{QE}
\tsc{EP}
\tsc{PMS}
\tsc{BEC}
\tsc{DE}
\begin{document}
    
    %
\let\WriteBookmarks\relax
\def\floatpagepagefraction{1}
\def\textpagefraction{.001}

\shorttitle{Estimation of the angular position of a two--wheeled balancing robot}

\shortauthors{K. Laddach et~al.}

\title[mode=title]{Estimation of the angular position of a two--wheeled balancing robot using a real IMU with selected filters}

\author[1]{Krzysztof Laddach}[orcid=0000-0001-9122-2167]%
\ead{krzysztof.laddach@pg.edu.pl}%

\author[1]{Rafa{\l} {\L}angowski}[orcid=0000-0003-1150-9753]%
\cormark[1]
\ead{rafal.langowski@pg.edu.pl}%

\author[1]{Tomasz Zubowicz}[orcid=0000-0002-0270-2474]%
\ead{tomasz.zubowicz@pg.edu.pl}%

\address[1]{Department of Electrical Engineering, Control Systems and Informatics\\Gda{\'n}sk University of Technology, ul. G. Narutowicza 11/12, 80-233 Gda{\'n}sk, Poland}%

\cortext[cor1]{Corresponding author}%
    \begin{abstract}
A low--cost measurement system using filtering of measurements for two--wheeled balancing robot stabilisation purposes has been addressed in this paper. In particular, a measurement system based on gyroscope, accelerometer, and encoder has been considered. The measurements have been corrected for deterministic disturbances and then filtered with Kalman, $\alpha$--$\beta$ type, and complementary filters. A quantitative assessment of selected filters has been given. As a result, the complete structure of a measurement system has been obtained. The performance of the proposed measurement system has been validated experimentally by using a dedicated research rig.%
\end{abstract}%

    \maketitle
    
    \section{Introduction}\label{sec:introduction}%
A two--wheeled balancing robot is a single--axle mobile vehicle with a centre of mass located above the wheel rotation axis enabling tilt in only one axis \cite{Li:2013}. This type of construction has been gaining popularity in recent years, especially in the field of commercial applications, which include, e.g., segway, hoverboard, etc. The basic functionality (the main control goal) of a two--wheeled balancing robot is to enable its movement (control of linear velocity) while stabilising it -- keeping a robot in a vertical position (control of angular position) \cite{Li:2013,Pratama:2015,Zhuang:2014,Mahmoud:2017}. Fundamentally, the goal of a two--wheeled balancing robot control coincides with an inverted pendulum control problem. Thus, many of the approaches used in solving an inverted pendulum stabilisation problem are applicable to a two--wheeled balancing robot control, e.g., \cite{Andrzejewski:2019,Waszak:2020,Saleem:2020,Jain:2021,Kharola:2016,Seekhao:2020}.%

In order to enable the basic functionality of a two--wheeled balancing robot an adequate control system is needed. This control system, apart from the use of a suitable control technique, depends on the information provided by the measurement system (measuring devices) and requires a properly selected actuator system, typically electric drives. Hence, one of the crucial factors which should be considered during control system design is the availability and quality of measurement information. Thus, this paper focuses on measuring issues in a control feedback loop for two--wheeled balancing robot stabilisation purposes.%

To accomplish the stabilisation control objective, at least an angular position of a two--wheeled balancing robot, i.e. the value of the angle of tilt of a two--wheeled balancing robot from the vertical axis must be known. This information is provided to a feedback loop by the measurement system. The measurement system can be interpreted as an assemblage of (hard) sensors, and (optionally) algorithms used to enhance or augment the information provided, e.g., by applying filters or estimators (soft--sensors) \cite{Maldonado:2019,Zhao:2018,Langowski:2017,Witkowska:2020}. Typically, a measurement system that provides high--quality measurements is an expensive investment. This issue is particularly important in the case of mentioned commercial applications. A typical solution to this problem is based on using a cheaper or fewer number of sensors. Unfortunately, in both cases, the resulting quality of provided measurements can be significantly decreased and in consequence insufficient for control purposes. Clearly, the measurement information can be influenced by measurement noise or errors. The widespread approach to cope with this problem is based on the use of estimation (filtration). Hence, by exploiting the soft--sensor approach it is possible to find a trade--off between the cost of a measurement system and the quality of measurement information. Thus, in this paper, a low--cost measurement system using filtering of measurements for two--wheeled balancing robot stabilisation purposes is further considered.%

In the considered application domain, typically an Inertial Measurement Unit (IMU) is a core of the measurement system. An IMU usually is composed of various configurations of gyroscopes, accelerometers, and magnetometers. In this paper, the IMU consists of gyroscopes and accelerometers. Thus, the required information regarding the angular position of a two--wheeled balancing robot is determined based on measurements from the gyroscopes and accelerometers. The gyroscope delivers measurements of the angular velocity of a two--wheeled balancing robot. Moreover, these measurements can be integrated in time, assuming the knowledge of the initial conditions, which allows computing the gyroscope's orientation \cite{Bortz:1971,Ignagni:1990,Barshan:1995,Ojeda:2002}. The precision gyroscopes, e.g., those based on optical phenomena such as ring laser \cite{Titterton:2004}, are too expensive and bulky for aforementioned applications including two--wheeled balancing robots. Therefore, the cheaper gyroscopes of the type Micro Electrical Mechanical Systems (MEMSs) are commonly used \cite{Yazdi:1998,Titterton:2004}. Besides its low purchase and operating costs, MEMS is characterised by a solid construction with small size and weight. In addition, these sensors are low power consumption, short start--up time, and high--reliability. Moreover, MEMS gyroscopes and accelerometers are capable of providing inertial--grade measurements of angular velocity and acceleration even for long--range navigation systems \cite{Titterton:2004}. However, these are also characterised by a lower quality of the measurements provided. Also, the integration of measurement errors leads to accumulating an error in the calculated orientation, which prevents the proper measurement of the absolute orientation of gyroscope \cite{Madgwick:2010}. Hence, an additional accelerometer (or magnetometer) is required to measure a gravitational (respectively magnetic) field vector with known orientation in space. However, these measurements are subject to the interference of accelerometer non-ideal characteristics (or imprecision) and a two--wheeled balancing robot vector of motion. Therefore, besides the appropriate collection of measurements from considered sensors, the filtration and correction of measurements are necessary \cite{Vaganay:1993}. The aim of filtration is to remove stochastic measurements interference whereas the purpose of correction is to improve measurements by removing deterministic interference. Hence, correction of measurements consists of taking into account the accelerations resulting from sensor movement, and identifiable measurement errors such as non-linearity or bias \cite{Vaganay:1993}. In the literature, various models of stochastic interference can be found; nevertheless, the most widespread seems to be Gauss--Markov model \cite{Park:2004,Park:2008,Park:2006,Nassar:2004}. Naturally, this also involves an extensive volume of literature on filtration \cite{Nassar:2004,Kalman:1960,Premerlani:2009,Vaganay:1993,Mahony:2005,Saho:2015,Madgwick:2010,Baldwin:2007,Mahony:2012}. The most common types of filters used in this task are: Kalman filter \cite{Kalman:1960,Lee:2012}, $\alpha$--$\beta$ filter and its extensions \cite{Saho:2015}, and complementary filter \cite{Mahony:2005,Lee:2012}. The popularity of complementary and $\alpha$--$\beta$ filters is due to their simplicity and computational efficiency, which translates into their performance and reduced need for microprocessor power. Whereas, Kalman filter provides the optimal estimates of the angle of a two--wheeled balancing robot tilt from the vertical axis, but only if certain assumptions are met \cite{Kalman:1960,Titterton:2004}. In turn, works in the correction of interference of measurements made by accelerometer and gyroscope can be found in, e.g., \cite{Park:2004,Park:2006,Park:2008,Skog:2006,Jiancheng:2009,Ramalingam:2009,Lee:2011,Mohammed:2018,Allen:1998}.%

In this paper, correction of measurements uses an additional measurement of a two--wheeled balancing robot's (progressive) linear position provided by the encoder. It should be noticed that, in the case of balancing robots, this does not introduce an additional cost for the measurement system, because most control systems use this information anyway. Whereas for the filtering of the measurements the following are used: Kalman filter, a family of $\alpha$--$\beta$ filters, and complementary filter. Thus, a novelty presented in the paper is a comparison of the use of correction and distinct filtration of measurements mechanisms in a single measurement system. Hence, the main contribution of this paper is to investigate a low--cost measurement system with correction and filtration of provided measurements for stabilisation of a two--wheeled balancing robot purposes. Moreover, a performance comparison of selected filters including attention to the `miss-use' of the `classical' Kalman filter is presented. Each filter has been designed and then implemented in the constructed two--wheeled balancing robot. To that goal, optimised (minimum covariance) infinite impulse response filters have been put against a minimum covariance, linear, and unbiased filtering implemented using recursive Kalman filter. In the case of the former solutions, this involves $\alpha$--$\beta$ type filters and complementary filter. The obtained results have been quantitatively assessed using a typical measure, i.e. mean square error ($\mathrm{MSE}$).%

The paper is organised as follows. The problem statement is presented in section \ref{sec:problem}. Section \ref{sec:measuring_devices} includes the description of the measurement system. Next, the experimental framework the results obtained are widely discussed in section \ref{sec:framework}. The paper is concluded in section \ref{sec:conclusions}.
    \section{Problem Statement}\label{sec:problem}%
\begin{figure}[!ht]
    \centering
    \includegraphics[width=0.75\columnwidth]{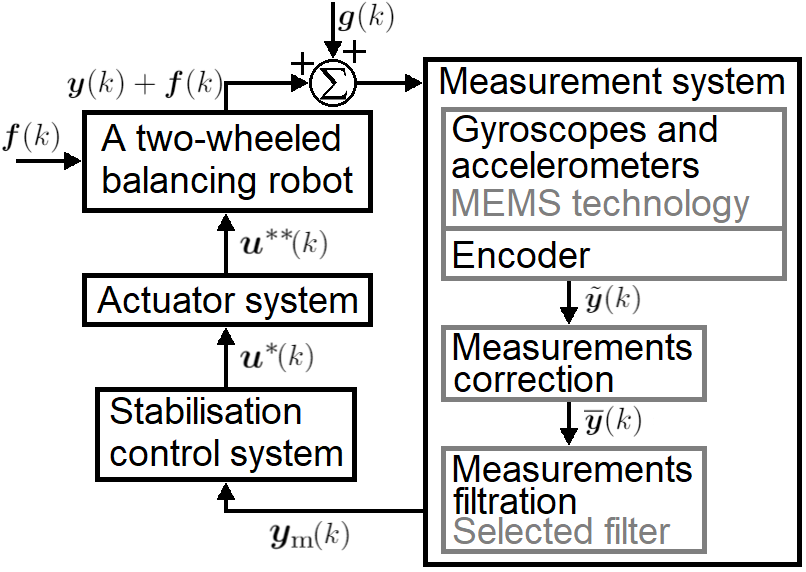}
    \caption{General structure of the two--wheeled balancing robot stabilisation control system}\label{fig:control_structure}
\end{figure}

Consider $\bm{y}$, $\tilde{\bm{y}}$, $\overline{\bm{y}}$, $\bm{y}_\mathrm{m}$, $\bm{u}^*$, and $\bm{u}^{**}$ to denote the vectors of: real measurements, sensors outputs, corrected measurements, measurements provided by the measurement system (after correction and filtration), control signals generated by the stabilisation control system, and control signals applied to a two--wheeled balancing robot, respectively. The general structure of a two--wheeled balancing robot stabilisation control system is shown in Fig.~\ref{fig:control_structure}. %

As it can be noticed in Fig.~\ref{fig:control_structure} the measurement system consists of two main parts. The first (physical/hardware layer) includes sensors, most notably gyroscopes and accelerometers. As it has been mentioned in section \ref{sec:introduction} these are made in MEMS technology. In turn, the second (software layer) includes correction and filtration mechanisms, primarily. The physical layer of the measurement system is described by the so--called measurement equation, which, in general form, yields:
\begin{equation}\label{eq:measur_equation}
      \tilde{\bm{y}}(k) = \bm{y}(k) + \bm{f}(k) + \bm{g}(k), 
\end{equation}
where: $k$ is the discrete time instant; $\bm{f}(k)$ represents deterministic measurements interference; $\bm{g}(k)$ represents stochastic measurements interference.

Thus, measurements $\tilde{\bm{y}}$ are assumed to be disturbed by deterministic and stochastic interference. This necessitates their correction and implies the existence of the second layer of measurement system. The software layer of the measurement system comprises correction and filtration. A detailed description of the measurements correction, which is based on the adopted model of deterministic interference, is presented in sections \ref{subsec:gyroscope} -- \ref{subsec:other_components}. Signal filtering, on the other hand, is performed using a selected filter, the design of which are presented in sections \ref{subsec:selected_filters}. Thus, as a result of the above operations, the measurement system will provide measurements $\bm{y}_\mathrm{m}$. It is easy to notice that in this case the vector $\bm{y}_\mathrm{m}$ is an estimate of the two--wheeled balancing robot's angular position, which is the necessary measurement information for the stabilisation control system.

The deliberation and conclusion provided in the following sections of the manuscript include have been obtained under the following conditions. For the purpose of estimation (filtration), the following model of two--wheeled balancing robot kinematics has been adopted \cite{Chhotray:2016}: 
\begin{equation} \label{eq:robot_kinematics}
    \begin{split}
        \phi(t) &= \phi_\mathrm{0} + \dot{\phi}(t)t + \frac{\ddot{\phi}(t)t^2}{2},\\
        \dot{\phi}(t) &= \dot{\phi}_\mathrm{0} + \ddot{\phi}(t)t,\\
        \ddot{\phi}(t) &= \ddot{\phi}(t),
    \end{split}
\end{equation}%
where: 
$t$ is the time instant; 
$\phi(t)$ denotes the angle of tilt of a two--wheeled balancing robot from the vertical axis (angular position) at time instant $t$;
$\dot{\phi}(t)$ signifies the angular velocity at time instant $t$;
$\ddot{\phi}(t)$ stands for the angular acceleration at time instant $t$;
$\phi_\mathrm{0}$ is the initial angular position of a two--wheeled balancing robot.

In this research work it is assumed that the angular position $\phi(\cdot)$ represent a real value of the angle of tilt of a two--wheeled balancing robot from the vertical axis.

Given \eqref{eq:measur_equation} and taking into account that the filtration system operates in discrete time, \eqref{eq:robot_kinematics} is discretised by the Euler method as follows:
\begin{equation} \label{eq:robot_kinematics_discr}
    \begin{split}
        \phi(k) &= \phi(k-1) + \dot\phi(k-1)\Delta t + \frac{\ddot{\phi}(k-1) \Delta t^2}{2},\\
        \dot\phi(k) &= \dot\phi(k-1) + \ddot{\phi}(k-1) \Delta t,\\
        \ddot\phi(k) &= \ddot\phi(k-1),
    \end{split}
\end{equation}
where $\Delta t$ is the discretisation time--step.

The equations \eqref{eq:robot_kinematics_discr} can be re--written in the vector--matrix general form as follows:
\begin{align}
        \bm{x}(k) &= \bm{A} \bm{x}(k-1) + \bm{B} \bm{u}(k-1) \label{rowDynamiki},\\
        \bm{y}(k) &= \bm{C} \bm{x}(k) \label{rowWyjscia}.
\end{align}

Equation \eqref{rowDynamiki} stands for the equation of system dynamics whereas \eqref{rowWyjscia} is the measurement (observation) equation. Hence, $\bm{x}(k)$, $\bm{u}(k-1)$, and $\bm{y}(k)$ represent the state vector of the system, input vector to the system, and output vector from the system, respectively, and they will change in a given considered filter. Similarly, the size and elements of matrices $\bm{A}$, $\bm{B}$, and $\bm{C}$ will change in a given considered filter.

As it has been aforementioned, the performance of selected filters, which consequently translates into the quality of the information provided by the measurement system, is quantitatively  assessed  using the mean square error of estimation, which is expressed by:
\begin{equation} \label{eq:MSE}
    \mathrm{MSE} = \frac{1}{m}\sum_{k=1}^{m}(\phi(k) - \hat{\phi}(k))^2, 
\end{equation}
where $\hat{(\cdot)}$ denotes the estimate of a given variable and $m$ signifies a number of discrete time instants.

To summarise, a low--cost measurement system with correction and filtration of provided measurements for stabilisation of a two--wheeled balancing robot is obtained. The entire measurement system has been implemented in the constructed two--wheeled balancing robot.

\begin{figure*}[!ht]
    \centering
    \includegraphics[width=0.8\textwidth]{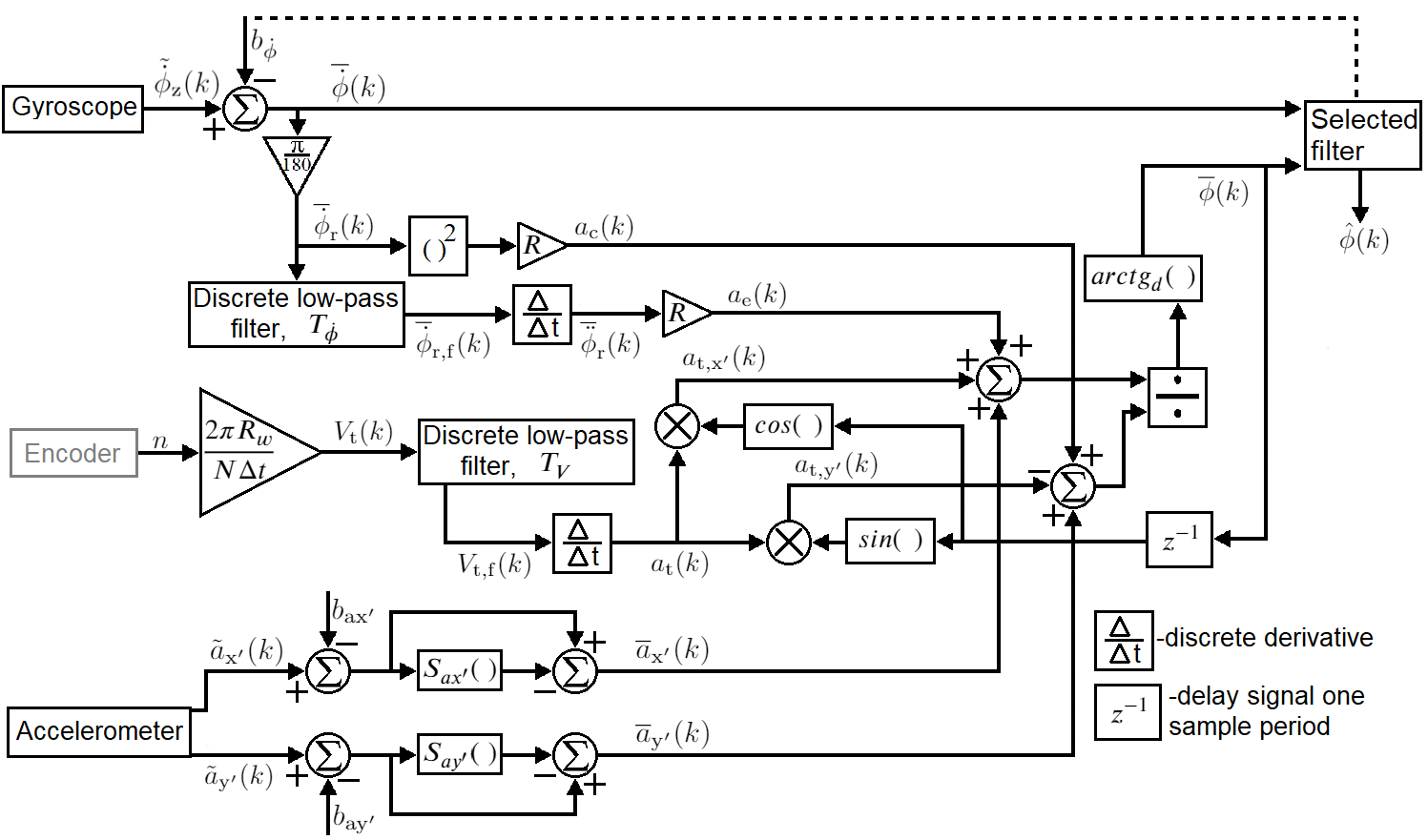}
    \caption{The detailed structure of the measurement system}\label{fig:struktura}
\end{figure*}%
    \section{Measurement system}\label{sec:measuring_devices}%
The detailed structure of the developed measurement system is shown in Fig. \ref{fig:struktura}. As it can be noticed, besides the aforementioned sensors, i.e. gyroscopes and accelerometers (MEMS), and encoder, and filters it also includes several other elements. Their detailed description is provided later in this section. As it has been mentioned, at the output of the measurement system there are estimates $\hat{\phi}(\cdot)$ of the angular position of the two--wheeled balancing robot.

\subsection{Gyroscope}\label{subsec:gyroscope}
\begin{figure}[!ht]
    \centering
    \includegraphics[width=0.85\columnwidth]{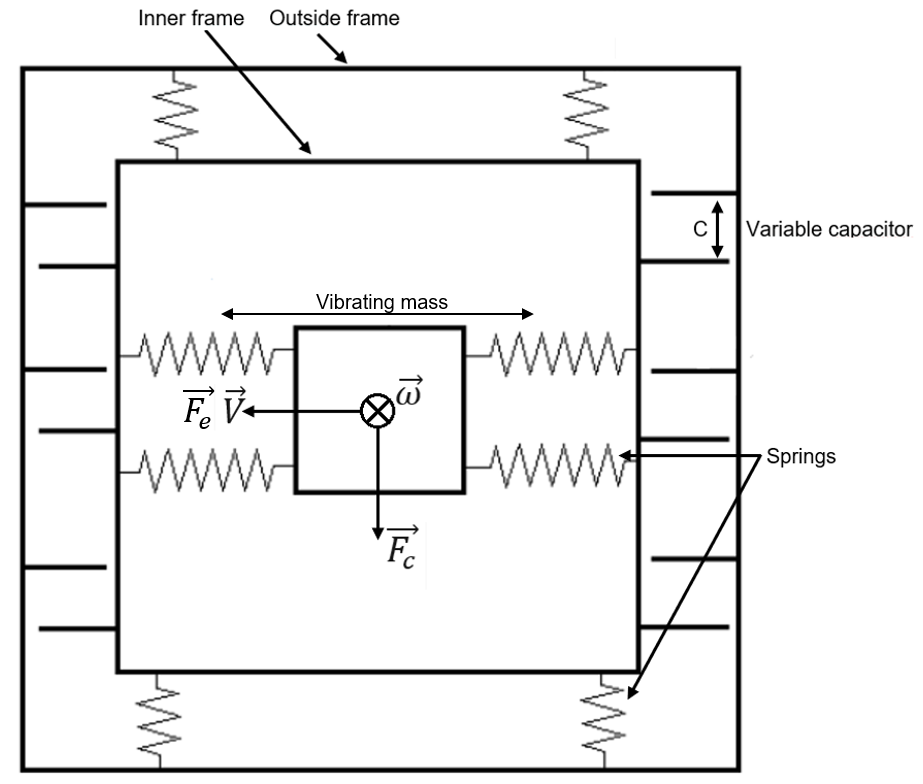}
    \caption{The internal structure of the MEMS gyroscope}\label{fig:budowaZyro}
\end{figure}

The simplified structure of the MEMS gyroscope is shown in Fig. \ref{fig:budowaZyro}. 
The particular symbols used in it denote the following: $C$ denotes the capacity, $\overrightarrow{F_\mathrm{e}}$ is the Euler force, $\overrightarrow{F_\mathrm{c}}$ denotes the centrifugal force, $\overrightarrow{V}$ signifies the horizontal velocity of vibrating mass, and $\overrightarrow{\omega}$ stands for angular velocity of the sensor. The measurements of the angular velocity of the two--wheeled balancing robot from this gyroscope are calculated by measuring the capacity $C$, which changes due to the motion of the inner frame relative to the outer frame. In the inner frame, the Coriolis force acts on the moving (oscillating, vibrating) proof mass, which causes movement of the inner frame. The inner frame can move only orthogonally to the direction of proof mass vibration. To eliminate the influence of inertia forces acting on the oscillating mass, which results from the gyroscope's progressive motion in the vibration direction, there are two sets of frames in the sensor in which mass vibrates in the anti-phase to each other. Then signals from both sets are added, and the in-phase component generated by linear acceleration is subtracted \cite{Fedorov:2015,Titterton:2004}. A more detailed description of the gyroscope's build and operation can be found in \cite{Titterton:2004,Park:2004}. 

According to \eqref{eq:measur_equation}, the equation for measuring angular velocity from a gyroscope in the selected axis can be written as follows \cite{Titterton:2004}:
\begin{equation}\label{eq:omega}
  \begin{split}
   \tilde{\dot{\phi}}_\mathrm{z}(k) 
    & = \dot{\phi}_\mathrm{z}(k) + S_{\dot\phi}\left(\dot{\phi}_\mathrm{z}(k)\right) + M_\mathrm{x} \dot{\phi}_\mathrm{x}(k) + M_\mathrm{y} \dot{\phi}_\mathrm{y}(k)\\
    & \quad{} + b_{\dot\phi} + B_\mathrm{z} a_\mathrm{z}(k) + B_\mathrm{y} a_\mathrm{y}(k) + B_\mathrm{zy} a_\mathrm{z}(k) a_\mathrm{y}(k)\\ 
    & \quad{} + \nu_{\dot\phi}, 
 \end{split}
\end{equation}
where: 
$\tilde{\dot{\phi}}_\mathrm{z}(k)$ denotes the measured value of the angular velocity in $\mathrm{z}$ axis at time instant $k$;
$\dot{\phi}_\mathrm{z}(k)$, $\dot{\phi}_\mathrm{x}(k)$, $\dot{\phi}_\mathrm{y}(k)$ are the real values of angular velocities in $\mathrm{z}$, $\mathrm{x}$, and $\mathrm{y}$ axes at time instant $k$, respectively;
$a_\mathrm{z}(k)$, $a_\mathrm{y}(k)$ stand for the accelerations in $\mathrm{z}$ and $\mathrm{y}$ axes at time instant $k$, respectively;
$b_{\dot\phi}$ is acceleration--insensitive bias;
$B_\mathrm{z}$, $B_\mathrm{y}$ are acceleration--sensitive bias coefficients in $\mathrm{z}$ and $\mathrm{y}$ axes, respectively;
$B_\mathrm{zy}$ signifies the anisoelastic bias coefficient;
$M_\mathrm{x}$, $M_\mathrm{y}$ denotes the cross--coupling coefficients in $\mathrm{x}$ and $\mathrm{y}$ axes, respectively;
$\nu_{\dot\phi}$ is the zero--mean random bias; 
$S_{\dot\phi}\left(\dot{\phi}_\mathrm{z}(k)\right)$ stands for the scale--factor error which may be expressed as a polynomial in $\dot{\phi}_\mathrm{z}(k)$ to represent the scale--factor non-linearities.

However, for non-pendulous designed MEMS type gyroscopes, which are built from three single axis gyroscopes reasonable is to expect that the cross--axis coupling factors and vibro-pendulous errors would be insignificant \cite{Park:2004,Park:2006,Park:2008}. Moreover, the scale--factor error arises mainly from temperature changes, and the resulting changes in the characteristics of the magnetic materials in the sensor \cite{Titterton:2004}. Assuming that the constructed two--wheeled balancing robot will move at a relatively constant temperature it is possible to neglect the scale--factor error. Thus, \eqref{eq:omega} can be re--written as follows:
\begin{equation}\label{eq:omega_simp}
    \tilde{\dot{\phi}}_\mathrm{z}(k) = \dot{\phi}_\mathrm{z}(k) + b_{\dot\phi} + \nu_{\dot\phi}. 
\end{equation}

Because in \eqref{eq:omega_simp} no elements are corresponding to phenomena in the other axes than $\mathrm{z}$ axis, in the further part of this paper the index of the $\mathrm{z}$ axis is omitted.

The bias $b_{\dot\phi}$ is non--zero value of the gyroscope's output even despite a lack of applied input rotation. It may be caused by various effects including the residual torques from flexible leads within the gyroscope, spurious magnetic fields, and temperature gradients. Moreover, this bias is independent of both rotational and progressive sensor movement \cite{Titterton:2004}. Thus, commonly the dynamics of $b_{\dot\phi}$ is assumed to zero (the value of bias is assumed to be constant) \cite{Titterton:2004,Park:2004,Park:2006,Park:2008}. Hence, the corrected measurements $\overline{\dot{\phi}}(k)$ of the angular velocity are calculated from:
\begin{equation}\label{eq:omega_uproszczone}
    \overline{\dot{\phi}}(k) = \tilde{\dot{\phi}}(k) - b_{\dot\phi}.
\end{equation}

\subsection{Accelerometer}\label{subsec:accelerometer}
The MEMS accelerometer belongs to the group of capacitive accelerometers. The principle of operation of capacitive accelerometers is analogous to the MEMS gyroscopes (see section~\ref{subsec:gyroscope}), except that the proof mass motion is caused directly by the inertia force, and the proof mass is not vibrated by electrical forces. A more detailed description can be found in \cite{Titterton:2004,Park:2004,Mohammed:2018}.

\begin{figure}[!ht]
    \centering
    \includegraphics[width=0.75\columnwidth]{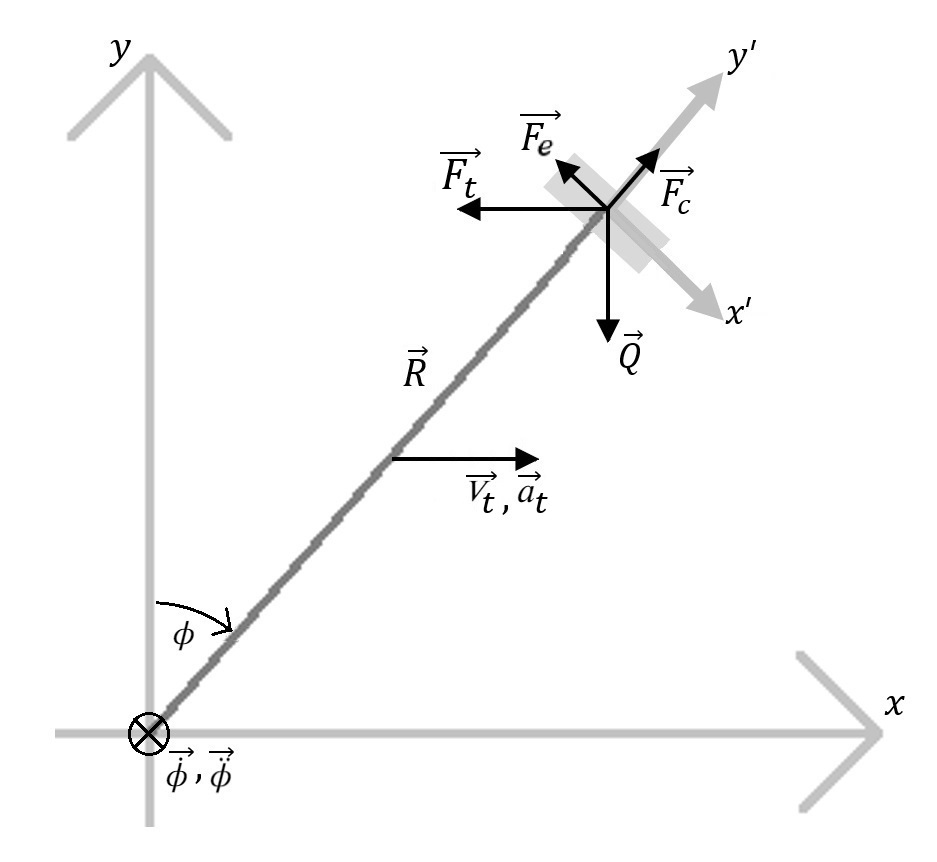}
    \caption{Distribution of forces on a moving sensor}\label{fig:accelerometer_axes}
\end{figure}

In the task of determining the value of the angle of tilt of the two--wheeled balancing robot from the vertical axis on a two--dimensional plane, from accelerometer the measurements of accelerations associated with the two axes, i.e. $\mathrm{x}'$ and $\mathrm{y}'$ are sufficient. Clearly, to calculate the angle of tilt, it is necessary to know the length of the component vectors of gravity acceleration (associated with the force of gravity $\overrightarrow{Q}$), which are projected on the axes of the two--wheeled balancing robot's reference system ($\mathrm{x}'$ and $\mathrm{y}'$). However, in addition to measuring the components of gravitational acceleration, the accelerometer measures also accelerations resulting from Coriolis, inertial $\overrightarrow{F_t}$, Euler $\overrightarrow{F_e}$ and centrifugal $\overrightarrow{F_c}$ forces. It is assumed that the velocities of the proof mass occurring in the accelerometer are so small and their duration is so short (the proof mass stabilises in the position where the applied forces cancel each other) that the Coriolis force acting on the proof mass is omitted from the consideration. The rest of the forces are shown in Fig. \ref{fig:accelerometer_axes}. Whereas the other symbols, i.e. $\overrightarrow{R}$, $\overrightarrow{V_t}$ and $\overrightarrow{a_t}$ stand for position vector, translational (in x axis) velocity and acceleration of the two--wheeled balancing robot, respectively.

According to \eqref{eq:measur_equation}, the equation for measuring acceleration from a accelerometer in the selected axis can be written as follows \cite{Titterton:2004}:
\begin{equation}\label{eq:accErrorModel}
  \begin{split}
    \tilde{a}_\mathrm{x'}(k) 
    & = a_\mathrm{x'}(k) + S_\mathrm{ax'}\left(a_\mathrm{x'}(k)\right) + M_\mathrm{y} a_\mathrm{y'}(k) + M_\mathrm{z} a_\mathrm{z'}(k)\\
    & \quad{} + b_\mathrm{ax'} + b_\mathrm{v} a_\mathrm{x'}(k) a_\mathrm{y'}(k) + \nu_\mathrm{ax'},
 \end{split}
\end{equation}
where:
$\tilde{a}_\mathrm{x'}(k)$ is the measured value of the acceleration in $\mathrm{x'}$ axis at time instant $k$;
$a_\mathrm{x'}(k)$, $a_\mathrm{y'}(k)$, $a_\mathrm{z'}(k)$ denote the real values of accelerations in $\mathrm{x'}$, $\mathrm{y'}$, and $\mathrm{z'}$ axes at time instant $k$, respectively;
$S_\mathrm{ax'}\left(a_\mathrm{x'}(k)\right)$ signifies the scale--factor error, usually expressed in polynomial form to include non-linear effects;
$M_\mathrm{y}$, $M_\mathrm{z}$ are the cross--coupling coefficients in $\mathrm{y}$ and $\mathrm{z}$ axes, respectively;  
$b_\mathrm{ax'}$ denotes the constant bias;
$b_\mathrm{v}$ stands for the vibro--pendulous error coefficient;
$\nu_\mathrm{ax'}$ is the random noise which expected value is assumed to be zero (see \cite{Titterton:2004}).

However, as for the gyroscope, also for accelerometer based on MEMS technology, non-pendulous design and consisting of three single accelerometers (one in each axis) the cross--axis coupling factors and vibro--pendulous errors would be insignificant \cite{Park:2004,Park:2006,Park:2008,Allen:1998}. Thus, \eqref{eq:accErrorModel} can be re--written as follows:  
\begin{equation}\label{eq:accErrorModel_corr}
    \tilde{a}_\mathrm{x'}(k) = a_\mathrm{x'}(k) + S_\mathrm{ax'}\left(a_\mathrm{x'}(k)\right) + b_\mathrm{ax'} + \nu_\mathrm{ax'}.
\end{equation}

Next, to compensate a calibration must be made to provide values of $b_\mathrm{ax'}$ and $S_\mathrm{ax'}\left(a_\mathrm{x'}(k)\right)$. A detailed description of this operation is contained in section \ref{sec:framework}. Hence, the corrected measurements $\overline{a}_\mathrm{x'}(k)$ of the acceleration in $\mathrm{x'}$ axis are calculated from:
\begin{equation}\label{eq:accErrorModel_corr_x}
    \overline{a}_\mathrm{x'}(k) = \tilde{a}_\mathrm{x'}(k) - b_\mathrm{ax'} - S_\mathrm{ax'}\left(\tilde{a}_\mathrm{x'}(k) - b_\mathrm{ax'}\right),
\end{equation}
and similarly in the $\mathrm{y'}$ axis from:
\begin{equation}\label{eq:accErrorModel_corr_y}
    \overline{a}_\mathrm{y'}(k) = \tilde{a}_\mathrm{y'}(k) - b_\mathrm{ay'}- S_\mathrm{ay'}\left(\tilde{a}_\mathrm{y'}(k) - b_\mathrm{ay'}\right).
\end{equation}

\subsection{Other components}\label{subsec:other_components}
According to Fig.~\ref{fig:struktura} the corrected value of the angle of tilt of the two--wheeled balancing robot from the vertical axis (the corrected angular position) can be determined as follows:
\begin{equation}\label{eq:arctg}
    \overline{\phi}(k) = \mathrm{arctg}_\mathrm{d} \left(\frac
        {\overline{a}_\mathrm{x'}(k) + a_\mathrm{e}(k) + a_\mathrm{t,x'}(k)}
        {\overline{a}_\mathrm{y'}(k) + a_\mathrm{c}(k) - a_\mathrm{t,y'}(k)}
    \right),
\end{equation}
where: 
$\overline{\phi}(k)$ is the corrected angular position at time instant $k$ in $\mathrm{[^\circ]}$;
$\mathrm{arctg}_\mathrm{d}(\cdot)$ denotes the arc tangent function;
$a_\mathrm{e}(k)$ stands for the Euler acceleration associated with Euler force at time instant $k$ \cite{Feynman:1963};
$a_\mathrm{c}(k)$ signifies the centrifugal acceleration associated with centrifugal force at time instant $k$ \cite{Feynman:1963};
$a_\mathrm{t, x'}(k)$, $a_\mathrm{t, y'}(k)$ are the accelerations resulting from the translational acceleration $\overrightarrow{a_\mathrm{t}}$ of the two--wheeled balancing robot in $\mathrm{x'}$ and $\mathrm{y'}$ axes at time instant $k$, respectively. 

The value of the centrifugal acceleration can be calculated as follows:
\begin{equation}\label{eq:centrifugal_acceleration}
     a_\mathrm{c}(k) = (\overline{\dot{\phi}}_\mathrm{r}(k))^2 R,
\end{equation}
where:
$\overline{\dot{\phi}}_\mathrm{r}(k) = \frac{\pi}{180}\overline{\dot{\phi}}(k)$ is the corrected measurements $\overline{\dot{\phi}}(k)$ of the angular velocity at time instant $k$ in $\mathrm{[rad/s]}$;
$R = 0.135 ~\mathrm{m}$ denotes the distance of the sensor from the axis of rotation.

In turn, the value of the Euler acceleration can be determined as:
\begin{equation}\label{eq:Euler_acceleration}
    a_\mathrm{e}(k) = \overline{\ddot{\phi}}_\mathrm{r}(k) R.
\end{equation}
where $\overline{\ddot{\phi}}_\mathrm{r}(k)$ is the angular acceleration in $\mathrm{[rad/s^2]}$. The value of $\overline{\ddot{\phi}}_\mathrm{r}(k)$ is not measured; therefore, it is calculated as a discrete derivative of the corrected angular velocity $\overline{\dot{\phi}}_\mathrm{r}(k)$ (see Fig.~\ref{fig:struktura}). 

However, the measurements of the angular velocity are burden with measurement noise, which excludes calculation of the derivative. Thus, to determine $\overline{\ddot{\phi}}_\mathrm{r}(k)$, the value $\overline{\dot{\phi}}_\mathrm{r}(k)$ is first passed through the following discrete first order low--pass filter:
\begin{equation}\label{eq:dolnoprzepustowy}
    \overline{\dot{\phi}}_\mathrm{r,f}(k) = \frac{\overline{\dot{\phi}}_\mathrm{r}(k)\Delta t + \overline{\dot{\phi}}_\mathrm{r,f}(k-1)T_\mathrm{\dot\phi}}{\Delta t + T_\mathrm{\dot\phi}},
\end{equation}
where $T_\mathrm{\dot\phi}$ is time constant of the low--pass filter (see section \ref{sec:framework}). 

Hence, the value of $\overline{\ddot{\phi}}_\mathrm{r}(k)$ yields:
\begin{equation}\label{eq:sec_der_phi}
    \overline{\ddot{\phi}}_\mathrm{r}(k) = \frac{
    \overline{\dot{\phi}}_\mathrm{r,f}(k)-\overline{\dot{\phi}}_\mathrm{r,f}(k-1)} {\Delta t}.
\end{equation}

The values of the accelerations $a_\mathrm{t,x'}(k)$ and $a_\mathrm{t,y'}(k)$ are calculated from the acceleration of the two--wheeled balancing robot $a_\mathrm{t}(k)$. This acceleration is discretely calculated as a derivative of the translational velocity $V_\mathrm{t}(k)$. It is calculated from the measurements of the additional encoder mounted on the axis of the two--wheeled balancing robot's wheels as follows \cite{Incze:2010}:
\begin{equation}\label{eq:velocity}
    V_\mathrm{t}(k) = \frac{2 \pi R_\mathrm{w}}{N \Delta t} n,
\end{equation}
where: 
$R_\mathrm{w} = 0.0375 ~[\mathrm{m}]$ is the radius of the two--wheeled balancing robot wheel;
$n$ denotes the number of counted pulses from encoder;
$N$ stands for the number of signal changes per shaft rotation.

It should be added that the errors resulting from the encoder measurements are not considered in the paper. It is due to their values are much smaller than the error values of the other considered sensors. Moreover, it is assumed that the considered two--wheeled balancing robot control system is tuned well-enough to avoid wheel slippage. However, the measurements of the progressive displacement have step characteristics. Indeed, $n$ is the number of counted pulses that is always an integer. Calculation of the derivative from such signal will not bring useful information. Therefore, the velocity is first filtered through the following low--pass filter:
\begin{equation}\label{eq:filtrPredkosciPost}
    V_\mathrm{t,f}(k) = \frac{V_\mathrm{t}(k) \Delta t + V_\mathrm{t,f}(k-1) T_\mathrm{V}}{\Delta t + T_\mathrm{V}},
\end{equation}
where $T_\mathrm{V}$ is time constant of the low--pass filter (see section \ref{sec:framework}).  

Thus, the value of the translational acceleration is then calculated using a discrete Euler derivative:
\begin{equation}\label{eq:translational_acceleration}
    a_\mathrm{t}(k) =\frac{V_\mathrm{t,f}(k) - V_\mathrm{t,f}(k-1)}{\Delta t}.
\end{equation}

Knowing the value of the translational acceleration, projection of it on the axes of the reference system associated with the sensor must be done. To do so, the corrected angular position must be used. Since its current value is just calculated, the previous value is used as follows:
\begin{equation}
    \begin{split}
        a_\mathrm{t,x'}(k) &= a_\mathrm{t}(k) \mathrm{cos}\left(\overline{\phi}(k-1)\right), \\
        a_\mathrm{t,y'}(k) &= a_\mathrm{t}(k) \mathrm{sin}\left(\overline{\phi}(k-1)\right).
    \end{split}
\end{equation}

\subsection{Selected filters}\label{subsec:selected_filters}
\begin{figure}[!ht]
    \centering
    \includegraphics[scale=0.3]{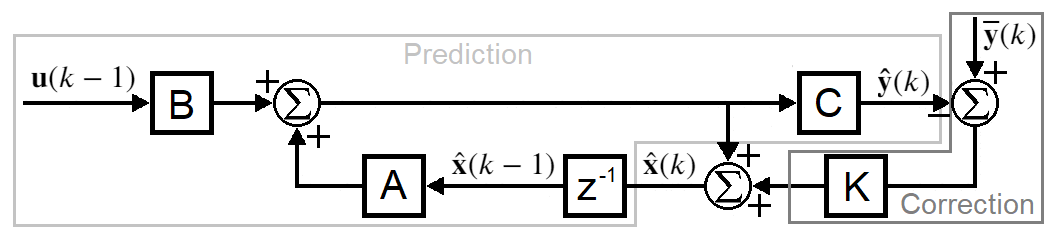}
    \caption{The structure of selected filters -- block `Selected filter' in Fig.~\ref{fig:struktura}}\label{fig:strukturaFiltru}
\end{figure}

As it has been aforementioned the Kalman filter, the family of $\alpha$--$\beta$ filters, and complementary filter have been used for the filtering (estimation) of the measurements. The structure of selected filters is shown in Fig.~\ref{fig:strukturaFiltru}.%

In turn, the dynamics of selected filter, in a general form, can be described by equation:
\begin{equation}\label{eq:dynamikaFiltru}
\begin{split}
    \hat{\bm{x}}(k) &= \left[\bm{A}-\bm{KCA}\right]\hat{\bm{x}}(k-1)\\
    & \quad{} + \left[\bm{B}-\bm{KCB}\right] \bm{u}(k-1) 
    + \bm{K}\overline{y}(k),
\end{split}
\end{equation}
where $\bm{K}$ is the gains matrix of the selected filter. 

Whereas the estimation error $\bm{e}(k)$ is defined as follows:
\begin{equation}\label{eq:estim_error}
    \bm{e}(k) = \bm{x}(k) - \hat{\bm{x}}(k).
\end{equation}

It is easy to show that the dynamics of the estimation error can be written as follows:  
\begin{equation}\label{eq:dynamikaBledu}
    \bm{e}(k) = \left[\bm{A}-\bm{KCA}\right]\bm{e}(k-1) + \bm{K} \left[\hat{\bm{y}}(k) - \overline{\bm{y}}(k)\right].
\end{equation}

To ensure the filter stability the following must hold:
\begin{equation}\label{eq:stability_condition}
   \forall_{\lambda_i \in \mathrm{eig}(\bm{A}-\bm{KCA})} \quad |\lambda_i| < 1,
\end{equation}
where:
$\lambda_i \in \mathrm{eig}(\bm{A}-\bm{KCA})$ is $i$th eigenvalue of matrix $[\bm{A}-\bm{KCA}]$;
$\mathrm{eig}(\bm{A}-\bm{KCA})$ denotes the spectrum of matrix $[\bm{A}-\bm{KCA}]$.

\subsubsection{Kalman filter}\label{kalman}
The filtration method, which ensures the optimal value of $\mathrm{MSE}$ of estimation for linear plants was developed by Rudolf E. Kalman in 1960 \cite{Kalman:1960}. In general, Kalman filter is a discreet observer in which the gain values of the correction part are calculated in each iteration of the filter in a way that results from the solution of the optimisation task. Assuming that the measurements are burdened only with Gauss noise this solution provides a minimal $\mathrm{MSE}$, in considered case between the corrected angular position and its estimate. It should be noticed that in fact, Kalman filter has no degrees of freedom, i.e. there are no adjustable parameters in it. The covariance matrices given when initialising the filter are derived from the interference characteristics and should be calculated from the Gauss noise analysis. However, this task provides many problems, especially when actual measurements are also affected by other types of interference, e.g. bias, or when the noise has not clear Gauss character, which often results in the empirical tuning of Kalman filter. A detailed description of Kalman filter can be found in \cite{Kalman:1960,Titterton:2004}. 

From the point of view of the above, two approaches to Kalman filter design are taken under consideration in this paper, `Kalman' and `Kalman$^*$'. The first assumes access to the a priori knowledge on the interference. The second assumes limited knowledge of the interference characteristics. Hence, in the first of these, the covariance matrices of the estimated variables $\bm{Q}$ and measurements $\bm{R}$, respectively are determined by using the interference analysis. In turn, in the second one (`Kalman$^*$'), the values of particular elements of the matrices $\bm{Q}$ and $\bm{R}$ have been selected by optimisation. It is decided that Kalman filter will provide estimates of the angular position $\hat{\phi}(k)$ and the gyroscope bias $\hat{b}_\mathrm{\dot\phi}$, whereas the corrected angular velocity measurement $\overline{\dot{\phi}}(k)$ is treated as an input to the kinematics model. This approach has been taken because it is most common for the task of filtering the angle of tilt of a two--wheeled balancing robot from the vertical axis. Moreover, \eqref{eq:robot_kinematics_discr} is simplify by assuming that $\ddot\phi(k) = 0$, whereas the dynamics of bias $b_\mathrm{\dot{\phi}}$ is assumed to be zero. Thus, the prediction equations can be written as follows:
\begin{equation}\label{eq:pred_Kalman}
\begin{split}
    \hat{\bm{x}}(k|k-1) &= \bm{A}\hat{\bm{x}}(k-1) +\bm{B} \overline{\dot{\phi}}(k-1), \\
    \hat{\phi}(k|k-1) &= \bm{C} \hat{\bm{x}}(k),
\end{split}
\end{equation}
where:
\begin{equation}
    \begin{split}
       \hat{\bm{x}}(\cdot) = \begin{bmatrix}
           \hat{\phi}(\cdot)\\
           \hat{b}_\mathrm{\dot\phi}(\cdot)
         \end{bmatrix}, \,
    \bm{A} = \begin{bmatrix}
        1 & -\Delta t\\
        0 & 1
       \end{bmatrix}, \, 
    \bm{B} = 
    \begin{bmatrix}
        \Delta t \\
        0
    \end{bmatrix}, \,
    \bm{C} = 
    \begin{bmatrix}
    1 \\ 0
    \end{bmatrix}^\mathrm{T}.
\end{split}
\end{equation}

The gains matrix (Kalman matrix) is of the form $\bm{K} = \begin{bmatrix} k_\mathrm{1} & k_\mathrm{2} \end{bmatrix}^{T}$, and it is calculated in each cycle of filter. For this purpose, in the first step the covariance matrix $\bm{P}$ is calculated as follows:
\begin{equation}
    \bm{P}(k|k-1) = \bm{A} \bm{P}(k-1)\bm{A}^{-1}+\bm{Q},
\end{equation}
where:
\begin{equation}
    \bm{Q} = 
    \begin{bmatrix}
        q_\mathrm{1} \Delta t & 0\\
        0 & q_\mathrm{2}
    \end{bmatrix},
\end{equation}
and $q_\mathrm{1}$, $q_\mathrm{2}$ are variances of $\hat{\phi}(\cdot)$ and $\hat{b}_\mathrm{\dot\phi}(\cdot)$, respectively.

Next:
\begin{equation}
   \begin{split}
      \bm{K}(k) &=
    \begin{bmatrix}
        k_\mathrm{1}(k) \\
        k_\mathrm{2}(k)
    \end{bmatrix} \\ &= 
    \bm{P}(k|k-1)\bm{C}^\mathrm{T} 
    \left(\bm{C}\bm{P}(k|k-1)\bm{C}^\mathrm{T}+\bm{R}\right)^\mathrm{-1},
\end{split}
\end{equation}
where $\bm{R} = [r]$, and $r$ is variances of measurements. 

Then the vector of estimates and the covariance matrix are subject to the correction:
\begin{equation}
   \begin{split}
     \hat{\bm{x}}(k|k) &= \hat{\bm{x}}(k|k-1) + \bm{K}(k)\left(\overline{\phi}(k)-\hat{\phi}(k|k-1)\right),
    \\
    \bm{P}(k|k) &= \left(\bm{I}-\bm{K}(k) \bm{C} \right) \bm{P}(k|k-1),
\end{split}
\end{equation}
where $\bm{I}$ is an identity matrix of a proper size.

The initial values of elements of matrix $\bm{P}$ have been calculated in such a way that in the first iteration of the filter, the correction gains take the value of the optimal $\alpha$--$\beta$ -- WB (see section \ref{sec:a_b_filters}) filter parameters. This has done because of the similarity of assumed kinematic model used in this two filters.

\subsubsection{The family of $\alpha$--$\beta$ filters}\label{sec:a_b_filters}
The $\alpha$--$\beta$ filter is a particular example of Kalman filter where the (correction) gains matrix is fixed and calculated outside the filter algorithm. There are many types of $\alpha$--$\beta$ filters; however, they are based on the same principle: the prediction is based on the dynamic model equations, and next, the predicted state is updated by the correction based on measurements. In various types of $\alpha$--$\beta$ filters, in the tasks of estimating orientation, the difference is the adopted model of kinematics. One of the simplest is $\alpha$--$\beta$ filter providing estimates of the angular position and velocity. This filter is described as first in this section. 

\medskip
\textbf{i) $\alpha$--$\beta$ filter without velocity bias estimation ($\alpha$--$\beta$ -- WOB)}

In this filter, model \eqref{eq:robot_kinematics_discr} is simplify by assuming that $\ddot\phi(k) = 0$. Whereas the estimated variables are the angular position $\hat{\phi}(k)$ and angular velocity $\hat{\dot{\phi}}(k)$. Thus, the prediction equations takes form:
\begin{equation}
  \begin{split}
    \hat{\phi}(k|k-1) &= \hat{\phi}(k-1) + \hat{\dot{\phi}}(k-1) \Delta t,\\
    \hat{\dot{\phi}}(k|k-1) &= \hat{\dot{\phi}}(k-1).
\end{split}
\end{equation}

In turn, in the correction phase, predictions are corrected by the equations:
\begin{equation}
  \begin{split}
     \hat{\phi}(k|k) &= \hat{\phi}(k|k-1) + \alpha\left(\overline{\phi}(k) - \hat{\phi}(k|k-1)\right), \\
    \hat{\dot{\phi}}(k|k) &= \hat{\dot{\phi}}(k|k-1) + \frac{\beta}{\Delta t} \left( \overline{\dot{\phi}}(k) - \hat{\dot{\phi}}(k|k-1)\right).
\end{split}
\end{equation}

Once above equations are bound, equation \eqref{eq:dynamikaFiltru} can be obtained in which the vectors and matrices take form:
\begin{equation}
  \begin{split}
      \hat{\bm{x}}(k) = \begin{bmatrix}
            \hat{\phi}(k) \\
            \hat{\dot{\phi}}(k)
        \end{bmatrix}, \, 
          \bm{A} =
           \begin{bmatrix}
            1 & \Delta t \\
            0 & 1
           \end{bmatrix},\,  
    \bm{B} = 0, \,\\  
    \bm{C} =
        \begin{bmatrix}
            1 & 0 \\
            0 & 1
        \end{bmatrix}, \, 
    \bm{K} = 
        \begin{bmatrix}
            \alpha & 0 \\
            0 & \beta
        \end{bmatrix}.
\end{split}
\end{equation}

\medskip
\textbf{ii) $\alpha$--$\beta$ filter with velocity bias ($\alpha$--$\beta$ -- WB)}

In this filter, model \eqref{eq:robot_kinematics_discr} is simplify again by assuming that $\ddot\phi(k) = 0$. However, due to the not ideal constancy of the $b_\mathrm{\dot{\phi}}$ value over time, also an estimate of its value is calculated in this filter. For this purpose, the bias dynamic is assumed as $\dot{b}_\mathrm{\dot{\phi}}(t) = 0$ which in discrete time gives $b_\mathrm{\dot{\phi}}(k) = b_\mathrm{\dot{\phi}}(k-1)$. The estimate of bias is initialised with the value calculated in section \ref{sec:framework}. The direct measurement of the angular velocity is treated as an input to the model. Thus, the prediction equations take form:
\begin{equation}
  \begin{split}
     \hat{\phi}(k|k-1) &= \hat{\phi}(k-1) + \left(\tilde{\dot{\phi}}(k-1) - \hat{b}_\mathrm{\dot{\phi}}(k-1)\right)\Delta t,\\
    \hat{b}_\mathrm{\dot{\phi}}(k|k-1) &= \hat{b}_\mathrm{\dot{\phi}}(k-1).
\end{split}
\end{equation}

Whereas the correction phase equations take the form of:
\begin{equation}
  \begin{split}
    \hat{\phi}(k|k) &= \hat{\phi}(k|k-1) + \alpha\left(\overline{\phi}(k) - \hat{\phi}(k|k-1) \right),\\
    \hat{b}_\mathrm{\dot{\phi}}(k|k) &= \hat{b}_\mathrm{\dot{\phi}}(k|k-1) + \beta \left(\overline{\phi}(k) - \hat{\phi}(k|k-1) \right).
\end{split}
\end{equation}

The vectors and matrices of equation \eqref{eq:dynamikaFiltru} take form:
\begin{equation}
   \begin{split}
      \hat{\bm{x}}(k) =
        \begin{bmatrix}
            \hat{\phi}(k) \\
            \hat{b}_\mathrm{\dot{\phi}}(k)
        \end{bmatrix}, \, 
    \bm{A} =
        \begin{bmatrix}
            1 & -\Delta t \\
            0 & 1
        \end{bmatrix}, \, 
    \bm{B} = 
        \begin{bmatrix}
            \Delta t \\
            0
        \end{bmatrix}, \\  
    \bm{C} =
        \begin{bmatrix}
            1 & 0
        \end{bmatrix}, \, 
    \bm{K} = 
        \begin{bmatrix}
            \alpha \\
            \beta
        \end{bmatrix}.
\end{split}
\end{equation}

One of the ways of using the angular velocity measurement is to treat it as an input into kinematics equations, which is shown above. In another filter from the $\alpha$--$\beta$ family, the angular velocity measurement is used differently, which is described in the next section.

\medskip
\textbf{iii) $\alpha$--$\beta$--$\theta$--$\gamma$ filter}

Similar to the $\alpha$--$\beta$ filters described previously, also in this filter the kinematics is simplified by assuming $\ddot\phi(k) = 0$. Also, the bias is not estimated. In contrast, this time both measurement values are used only in the correction phase. Thus, the prediction phase consists of equations:
\begin{equation}
  \begin{split}
      \hat{\phi}(k|k-1) &= \hat{\phi}(k-1)+\Delta t
      \hat{\dot{\phi}}(k-1),\\
      \hat{\dot{\phi}}(k|k-1) &= \hat{\dot{\phi}}(k-1).
\end{split}
\end{equation}

Whereas, the correction phase is extended by using both measurements in both its equations:
\begin{equation}
  \begin{split}
    \hat{\phi}(k|k) 
    & = \hat{\phi}(k|k-1) 
    + \alpha\left(\overline{\phi}(k)-\hat{\phi}(k|k-1)\right)\\  
    & \quad{} + \theta\Delta t \left(\overline{\dot{\phi}}(k) - \hat{\dot{\phi}}(k|k-1)\right),\\
    \hat{\dot{\phi}}(k|k)
    & = \hat{\dot{\phi}}(k|k-1) + \frac{\beta}{\Delta t}\left(\overline{\phi}(k)- \hat{\phi} (k|k-1) \right)\\ 
    & \quad{} + \gamma\left(\overline{\dot{\phi}}(k)-\hat{\dot{\phi}}(k|k-1)\right).
\end{split}
\end{equation}

This filter fits into equation \eqref{eq:dynamikaFiltru} by the following vectors and matrices:
\begin{equation}
  \begin{split}
     \hat{\bm{x}}(k) =
        \begin{bmatrix}
            \hat{\phi}(k) \\
            \hat{\dot{\phi}}(k)
        \end{bmatrix}, \, 
    \bm{A} =
        \begin{bmatrix}
            1 & \Delta t \\
            0 & 1
        \end{bmatrix}, \,  
    \bm{B} = 0, \\  
    \bm{C} =
        \begin{bmatrix}
            1 & 0 \\
            0 & 1
        \end{bmatrix}, \, 
    \bm{K} = 
        \begin{bmatrix}
            \alpha & \theta \Delta t \\
            \frac{\beta}{\Delta t} & \gamma
        \end{bmatrix}.
\end{split}
\end{equation}

\medskip
\textbf{iv) $\alpha$--$\beta$--$\theta$ filter with acceleration ($\alpha$--$\beta$--$\theta$ -- WA-a or b)}

As it can be noticed in all filters presented above assumed that the angular acceleration is zero. In contrast in the last filter from the family, $\alpha$--$\beta$ described in this section does not do this, and the value of angular acceleration also is estimated. Thus, the prediction equations yields:
\begin{equation}
   \begin{split}
     \hat{\phi}(k|k-1) &= \hat{\phi}(k-1)
    +\Delta t \hat{\dot{\phi}}(k-1)
    +\frac{\Delta t^2}{2} \hat{\ddot{\phi}}(k-1),\\
    \hat{\dot{\phi}}(k|k-1) &= \hat{\dot{\phi}}(k-1)+\Delta t \hat{\ddot{\phi}}(k-1),\\
    \hat{ \ddot{\phi}}(k|k-1) &=  \hat{\ddot{\phi}}(k-1).
 \end{split}
\end{equation}

During the correction phase, the value of the angular acceleration can be updated based on the measurement of angular position or angular velocity \cite{Saho:2015}, what provides to the two following sets of equations, which are different only in the last one:
\begin{equation}
  \begin{split}
    \hat{\phi}(k|k) = \hat{\phi}(k|k-1) + \alpha \left(\overline{\phi}(k)-\hat{\phi}(k|k-1)\right),\\
    \hat{\dot{\phi}}(k|k) = \hat{\dot{\phi}}(k|k-1) + 
    \beta\left(\overline{\dot{\phi}}(k)-\hat{\dot{\phi}}(k|k-1)\right),\\
    iv.a)\ \hat{\ddot{\phi}}(k|k) = \hat{\ddot{\phi}}(k|k-1) + \frac{\theta}{\Delta t^2} \left(\overline{\phi}(k)-\phi(k|k-1) \right), \\
    iv.b)\ \hat{\ddot{\phi}}(k|k) = \hat{\ddot{\phi}}(k|k-1) + \frac{\theta}{\Delta t} \left(\overline{\dot{\phi}}(k)-\hat{\dot{\phi}}(k|k-1)\right).
\end{split}
\end{equation}

The above equations translate into appropriate vectors and matrices in equation \eqref{eq:dynamikaFiltru}:
\begin{equation}
  \begin{split}
    \hat{\bm{x}}(k) =
        \begin{bmatrix}
            \hat{\phi}(k) \\
            \hat{\dot{\phi}}(k) \\
            \hat{\ddot{\phi}}(k)
        \end{bmatrix}, \, 
    \bm{A} =
        \begin{bmatrix}
            1 & \Delta t & \frac{\Delta t^2}{2}\\
            0 & 1 & \Delta t \\
            0 & 0 & 1
        \end{bmatrix}, \\  
    \bm{B} = 0, \,  
    \bm{C} =
        \begin{bmatrix}
            1 & 0 & 0 \\
            0 & 1 & 0
        \end{bmatrix}, \\ 
    iv.a) \bm{K} = 
        \begin{bmatrix}
            \alpha & 0 \\
            0 & \beta \\
            \frac{\theta}{\Delta t^2} & 0
        \end{bmatrix}, \,
    iv.b) \bm{K} = 
        \begin{bmatrix}
            \alpha & 0 \\
            0 & \beta \\
            0 & \frac{\theta}{\Delta t}
        \end{bmatrix}.
\end{split}
\end{equation}

In following part of the paper both possibilities are analysed, calling them $\alpha$--$\beta$--$\theta$ -- WA-a and $\alpha$--$\beta$--$\theta$ -- WA-b, respectively.

\subsubsection{Complementary filter}
Complementary filters integrate information of the same type coming from different sources, e.g., the measurements of the same physical variable from two or more different sensors. Hence, the purpose of this filtration is to use useful information while rejecting interference from a given source. This filtration brings the expected results when the sensors differ in the nature of the interference, particularly in its frequency \cite{Grygiel:2014}. This requirement is met in the analysed example, where the measurements from the accelerometer are burden with high--frequency noise, whereas the measurements from the gyroscope are burden mainly with a slow changing (low--frequency) bias error \cite{Titterton:2004}. To obtain the value of the angular position $\overline{\phi}_\mathrm{\dot{\phi}}$ from the measurements of the angular velocity, it has to be integrated, what in discrete time takes as follows:
\begin{equation}\label{eq:dtComp}
    \overline{\phi}_\mathrm{\dot{\phi}}(k) = \overline{\phi}_\mathrm{\dot{\phi}}(k-1) + \Delta t \overline{\dot{\phi}}(k). 
\end{equation}

Then the measurements obtained from the gyroscope are subjected to high--pass filtration, whereas the measurements from the accelerometer are subjected to low--pass filtration. The complementary filter should meet the rule:
\begin{equation}\label{eq:tfSum}
    \sum_{j=1}^{l} G_{j}(s) = 1,
\end{equation}
where $G_{j}(s)$ is the transfer functions of $j$th sub--filter. 

There is an infinite number of pairs consisting of high--pass and low--pass filter that meets \eqref{eq:tfSum}. However, i.a., in order to ensure a low computational cost, the simple case has been selected, taking the low--pass filter as first--order inertia:
\begin{equation}\label{eq:tran_l_p}
    G_\mathrm{l-p}(s) = \frac{1}{1 + T_\mathrm{c} s},
\end{equation}
where $T_\mathrm{c}$ is the filter parameter. 

Thus, according to \eqref{eq:tran_l_p}, \eqref{eq:tfSum} requires that the transfer function of the high--pass filter is as follows:
\begin{equation}\label{eq:tran_h_p}
    G_\mathrm{h-p}(s) = \frac{T_\mathrm{c} s}{1+T_\mathrm{c} s}.
\end{equation}

The output from the complementary filter $\hat{\phi}(k)$ is the sum of output signals of the low--pass and high--pass filters. By using the inverse Laplace's transformation and discretising with the backward Euler method, this sum can be approximate as follows \cite{Grygiel:2014}:
\begin{equation}\label{eq:comp}
  \begin{split}
    \hat{\phi}(k) 
    & = \frac{T_\mathrm{c}}{\Delta t + T_\mathrm{c}}\hat{\phi}(k-1)
    + \frac{\Delta t}{\Delta t + T_\mathrm{c}}\overline{\phi}(k)\\
    & \quad{} + \frac{T_\mathrm{c}}{\Delta t + T_\mathrm{c}} \left(\overline{\phi}_\mathrm{\dot{\phi}}(k) -\overline{\phi}_\mathrm{\dot{\phi}}(k-1) \right).
  \end{split}    
\end{equation}

By inserting \eqref{eq:dtComp} into \eqref{eq:comp} it is obtained:
\begin{equation}\label{eq:compFinal}
\begin{split}
    \hat{\phi}(k) 
    & = \frac{T_\mathrm{c}}{\Delta t + T_\mathrm{c}}\hat{\phi}(k-1)
    + \frac{\Delta t}{\Delta t + T_\mathrm{c}}\overline{\phi}(k)\\
    & \quad{} + \frac{T_\mathrm{c}}{\Delta t + T_\mathrm{c}}\Delta t \overline{\dot{\phi}}(k).
 \end{split}    
\end{equation}

Equation \eqref{eq:compFinal} can be written as \eqref{eq:dynamikaFiltru} by using the following vectors and matrices:
\begin{equation}
  \begin{split}
    \hat{\bm{x}}(k) =
        \begin{bmatrix}
            \hat{\phi}(k)
        \end{bmatrix}, \, 
    \bm{A} =
        \begin{bmatrix}
            \frac{T_\mathrm{c}}{\Delta t + T_\mathrm{c}}
        \end{bmatrix}, \\  
    \bm{B} = 
        \begin{bmatrix}
            \frac{\Delta t}{\Delta t + T_\mathrm{c}} & \frac{T_\mathrm{c}\Delta t}{\Delta t + T_\mathrm{c}}
        \end{bmatrix}, \\  
    \bm{C} = 0, \, 
    \bm{K} = 0, \, 
    \bm{u}(k) = 
        \begin{bmatrix}
            \overline{\phi}(k) \\
            \overline{\dot{\phi}}(k)
        \end{bmatrix}.
\end{split}
\end{equation}

It should be noticed that the inputs $\bm{u}(k)$ are values from the current sample $k$ and not the previous one $k-1$.

\medskip
The developed a low--cost measurement system with correction and filtration of provided measurements is characterised by computational efficiency, and it is easy to automate the procedure of implementation in numerous devices of one type, where different values of errors can occur, e.g. gyroscope bias, due to non-identity of the applied sensors of a given type.
\section{Experimental framework and results} \label{sec:framework}
\begin{figure}[!ht]
    \centering
    \includegraphics[width=0.4\columnwidth]{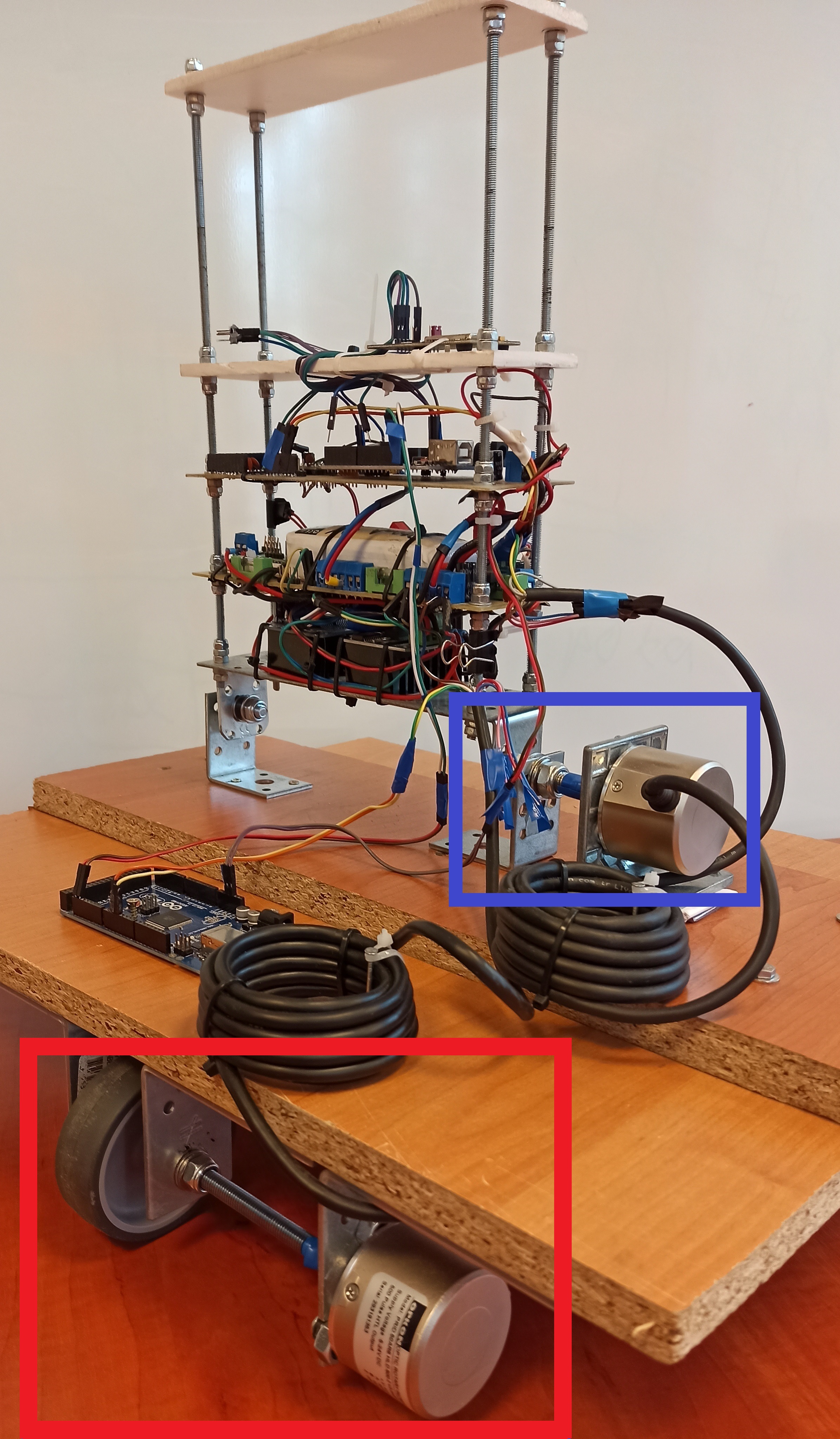}
    \caption{Research rig with two--wheeled balancing robot}\label{fig:research_stand}
\end{figure}

The research rig with constructed two--wheeled balancing robot is presented in Fig.~\ref{fig:research_stand}. 
It should be noticed that besides the aforementioned additional encoder (marked with a red box in Fig. \ref{fig:research_stand}) there is the second encoder (marked with a blue box in Fig. \ref{fig:research_stand}). This encoder is used to acquire reference measurement values of the angular position of the constructed two--wheeled balancing robot. In practice, this requires information of the initial value of the angle of tilt of the two--wheeled balancing robot from the vertical axis and then the subsequent addition of the measured values to the current position. In other words, it requires changes in the position of the reference encoder shaft connected to the two--wheeled balancing robot in the axis of rotation, which is described by the formula:
\begin{equation}
    \phi(k) = \phi(k-1)+\Delta \phi(k),
\end{equation}
where $\Delta \phi(k)$ is calculated from encoder measurements as $\Delta \phi(k) = 360\, n / N$, and $N = 2000$.

In turn, a zero initial value of $\phi(k=0)$ is ensured by positioning the two--wheeled balancing robot vertically at the beginning of the data acquisition process.

\begin{figure}[b]
    \centering
    \includegraphics[width=0.9\columnwidth]{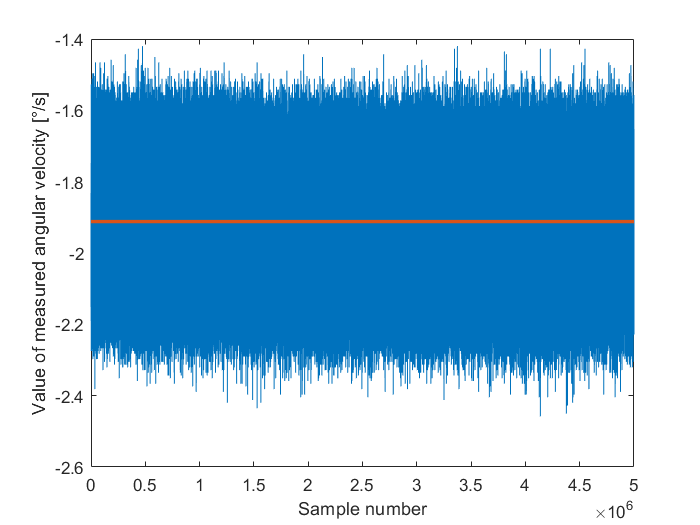}
    \caption{The measurements of angular velocity from gyroscope}\label{fig:biasGyro}
\end{figure}

\begin{figure}[t]
    \centering
    \includegraphics[width=0.9\columnwidth]{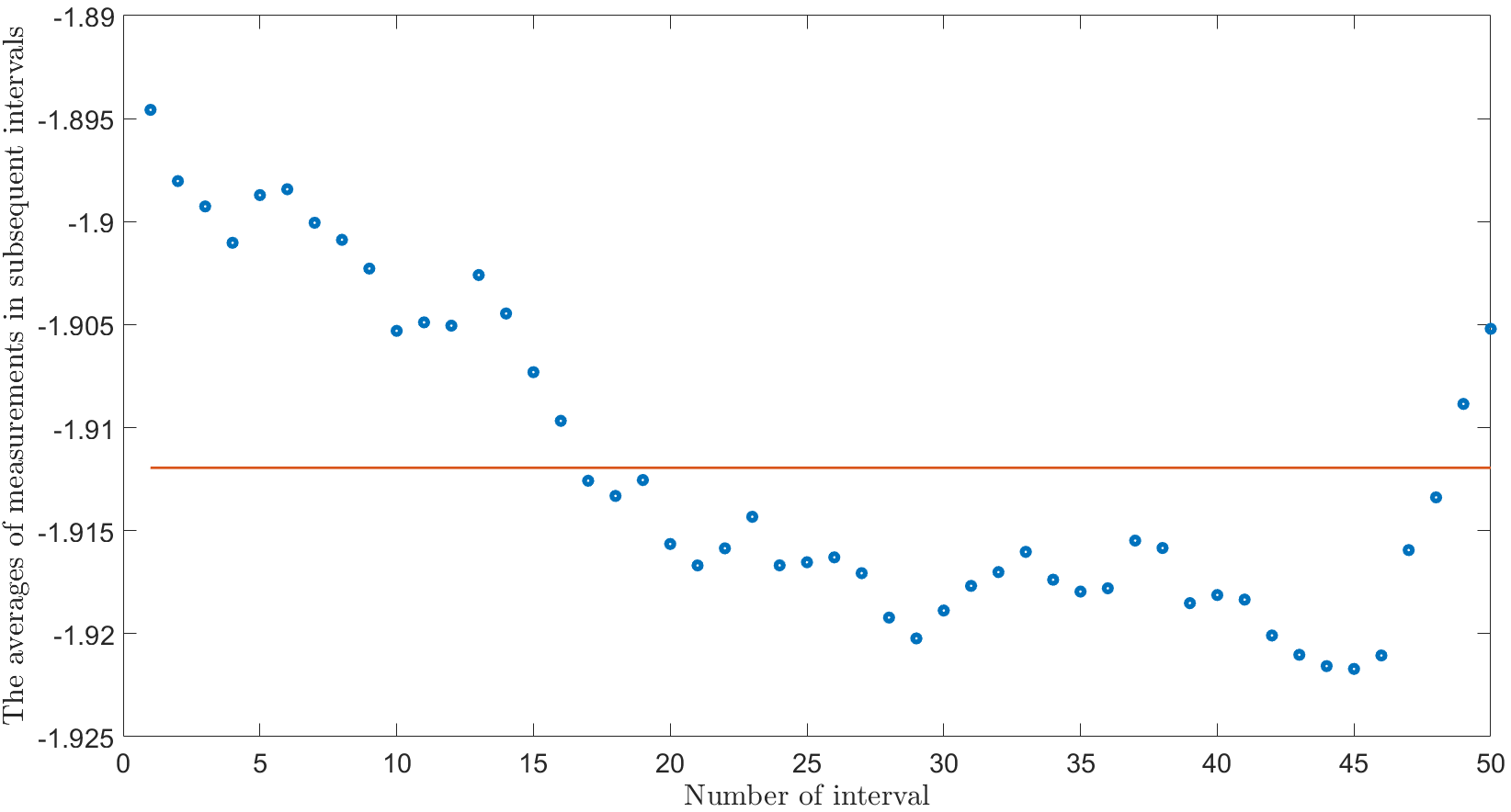}
    \caption{Mean values of the trajectory shown in Fig.~\ref{fig:biasGyro} in intervals of $10^{5}$ samples}\label{fig:co100tys}
\end{figure}

As the IMU, the MPU6050 unit has been chosen \cite{Invensense:2021}. This unit is equipped with three single--axis accelerometers and three single--axis gyroscopes in MEMS technology, which are set at right angles to each other. The configuration of the MPU6050 during the experiments has been as follows. The gyroscopes and accelerometers scale ranges have been set at $\pm250^\circ$ and $\pm2g$, respectively. The measurements have been read directly from the measurement registers without using a low--pass filter for the accelerometer and a high--pass filter for the gyroscope, which are built into the unit. The bias $b_\mathrm{\dot{\phi}}$ has been calculated as an average value of data from a two-hour range, in which the constructed two--wheeled balancing robot has been stationary. It has been checked if this approach is correct for the chosen sensor. For this purpose, with a sampling period equal $\Delta t = 1.508 ~\mathrm{ms}$, for about $2$ hours $5 \cdot 10^{6}$ measurements of the angular velocity have been collected -- see Fig. \ref{fig:biasGyro}. 
Then the minimum and maximum values have been read as $-1.41985 ~\mathrm{^\circ/s}$ and $-2.45802 ~\mathrm{^\circ/s}$, respectively. It should be noticed that some correlations can be found in changes of bias. More specifically, in Fig. \ref{fig:co100tys} the mean values of the trajectory from Fig. \ref{fig:biasGyro} in successive intervals, where each of all has $10^{5}$ samples are presented. Despite observable changes in the bias value over consecutive observation time windows (Fig. \ref{fig:co100tys}), its relative change over a long horizon remains small. Hence, although approaches that update the bias values, e.g., \cite{Park:2004,Park:2006,Park:2008} are justified, an approach based on single value bias estimation remains concurrent in the considered case. Hence, in this paper the value of bias $b_\mathrm{\dot{\phi}}$ has been established as $-1.91195 ~\mathrm{^\circ/s}$.

Similarly, the biases $b_\mathrm{ax'}$ and $b_\mathrm{ay'}$ have been calculated for the accelerometer. For this purpose, for about $2$ hours, the constructed two--wheeled balancing robot has been placed vertically stationary and measured the accelerations in two proper axes have been recorded. On the basis of the difference between the minimum values equal to $8.98132 ~\mathrm{m/s^2}$ and $-0.14610 ~\mathrm{m/s^2}$ for $\mathrm{y'}$ and $\mathrm{x'}$ axis, respectively, and the maximum values equal to $9.39087 ~\mathrm{m/s^2}$ and $0.14370 ~\mathrm{m/s^2}$ for $\mathrm{y'}$ and $\mathrm{x'}$ axis, respectively the correctness of a constant values of biases have been assumed (it should be mentioned that the measurements are also burden with stochastic interference). Finally, the biases $b_\mathrm{ax'}$ and $b_\mathrm{ay'}$ have been calculated as the mean value of data from a 2-hour range, and they are $-0.02340 ~\mathrm{m/s^2}$ and $-0.63629 ~\mathrm{m/s^2}$, respectively.
Then the search of the scale--factor errors -- $S_\mathrm{ay'}(\cdot)$ and $S_\mathrm{ax'}(\cdot)$ functions began. To find them the constructed two--wheeled balancing robot has been deflected from the vertical and stopped in various angles to collect the values of $\phi(\cdot)$. During this operation, the accelerations $\tilde{a}_\mathrm{y'}(\cdot)$ and $\tilde{a}_\mathrm{x'}(\cdot)$ have been also measured, in situations where have not been the accelerations associated with the sensor movement. On this basis, the reference accelerations have been defined as:
\begin{align*}
    a_\mathrm{ref,x'}(k) = g \sin\left(\phi(k)\right),\\
    a_\mathrm{ref,y'}(k) = g \cos\left(\phi(k)\right).
\end{align*}

\begin{figure*}[!ht]
    \centering
    \includegraphics[width=0.9\textwidth]{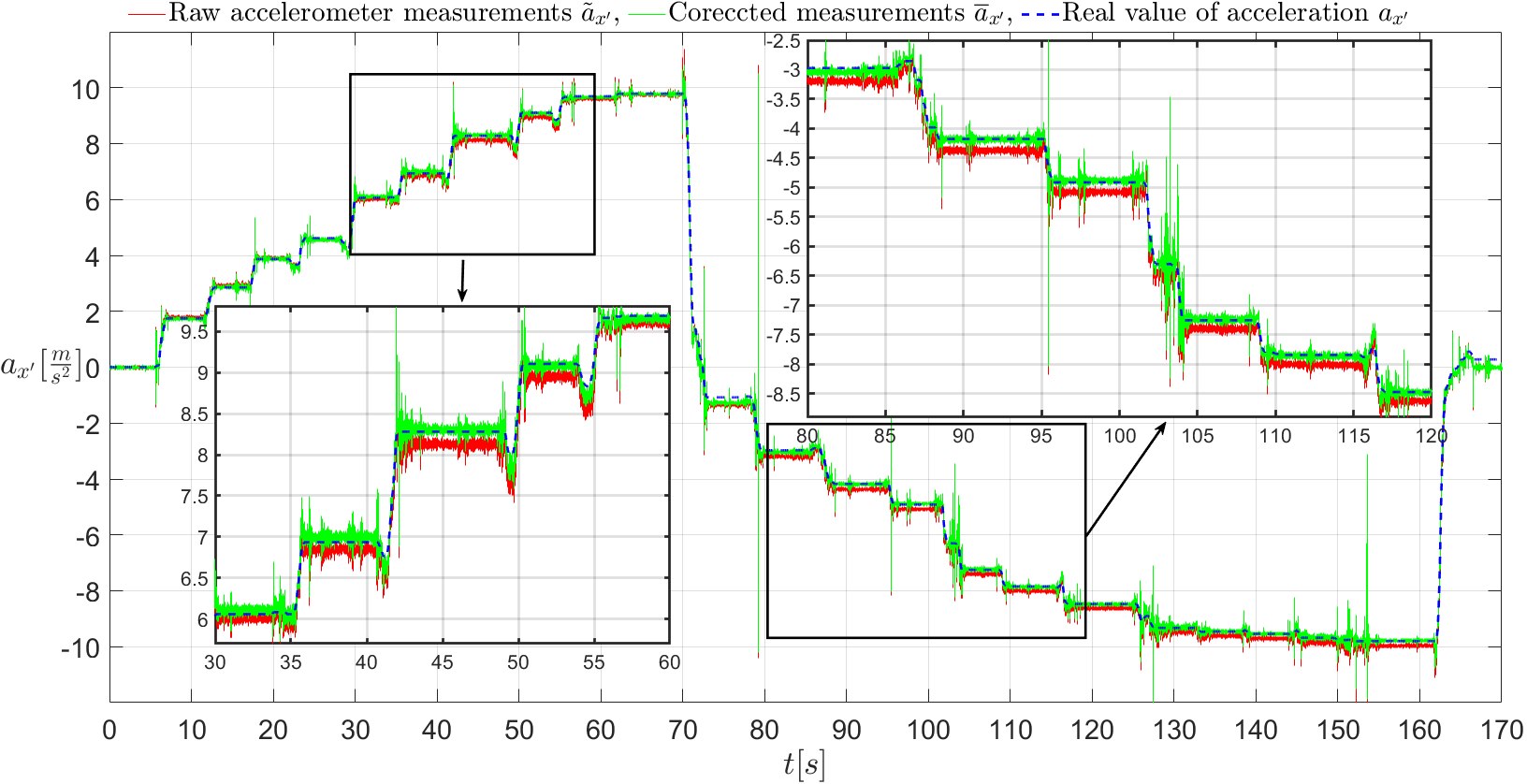}
    \caption{The trajectories of the acceleration in $\mathrm{x'}$ axis.}\label{fig:blad_nieliniowosciPOZIOM}
\end{figure*}

\begin{figure*}[!ht]
    \centering
    \includegraphics[width=0.9\textwidth]{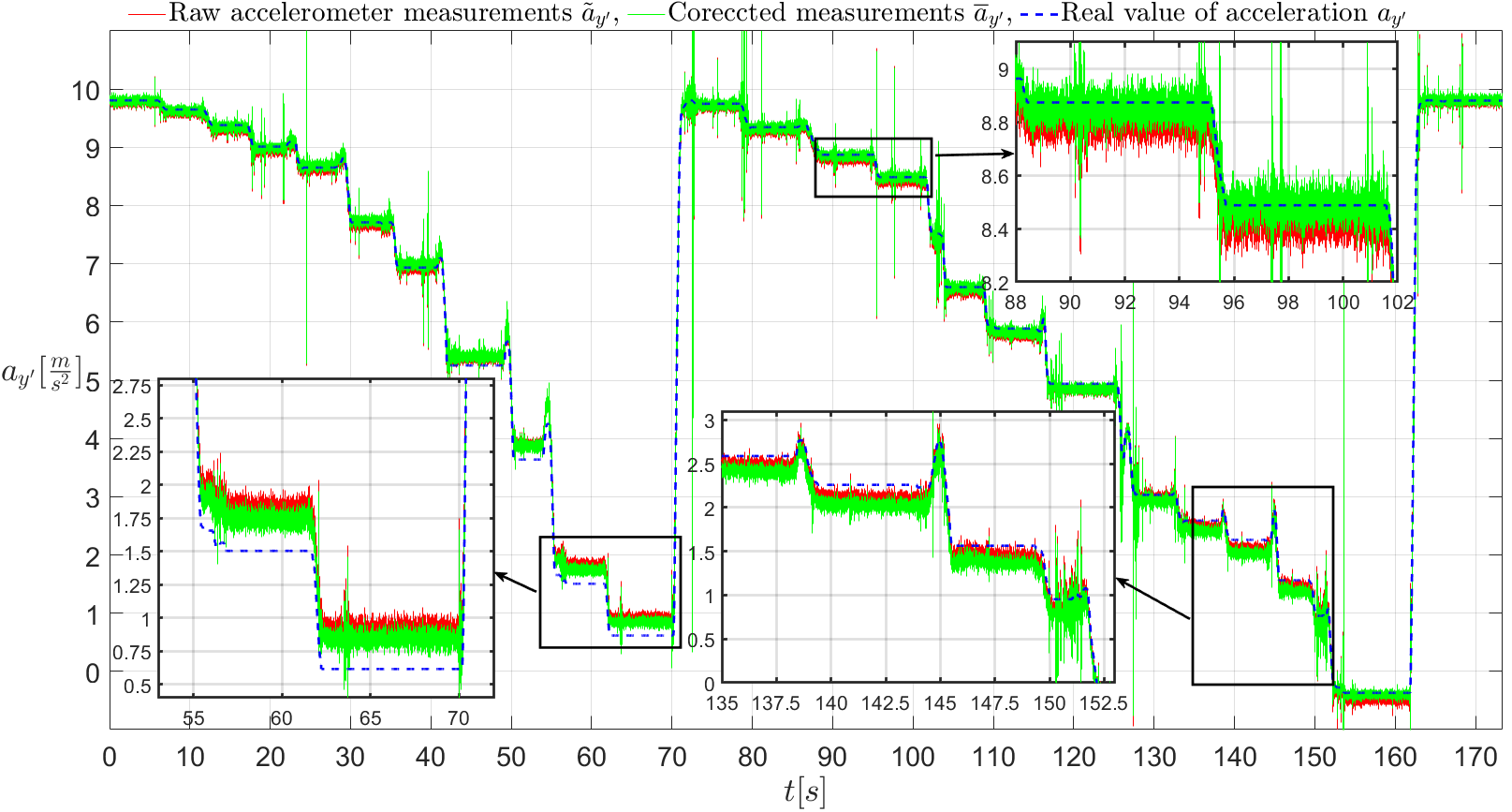}
    \caption{The trajectories of the acceleration in $\mathrm{y'}$ axis}\label{fig:blad_nieliniowosciPION}
\end{figure*}

Next, by solving an optimisation task that searches for function coefficients that ensure the minimum of $\mathrm{MSE}$ between the calculated corrected and reference acceleration, the parameters of searched functions are found. For both axes, polynomials from 1 to 10 degree have been tested. It has been observed that increase the polynomial degree above 5 brought degradation in the target function value. In all polynomials, free expression equal to zero has been assumed, because this value corresponds to the bias \cite{Titterton:2004}. 

The scale--factor error depends on the real acceleration of the sensor, therefore, the argument of the proposed functions is the measurement of acceleration corrected by bias, i.e. $\tilde{a}_\mathrm{x' \, or \, y'}(k)$ - $b_\mathrm{ax' \, or \, ay'}$. In other words, the measurements of acceleration are first corrected by the bias. Finally, functions $S_\mathrm{ay'}(\cdot)$ and $S_\mathrm{ax'}(\cdot)$ achieving the smallest $\mathrm{MSE}$ have been selected (optimisation has been carried out using \texttt{fminsearch} in MATLAB environment), which are of the form:

\begin{equation}
\begin{split}
    \mathit{Let:}\quad{} \, p_\mathrm{x}(k) 
    &:=  \tilde{a}_\mathrm{x'}(k) - b_\mathrm{ax'},\\
    S_\mathrm{ax'}\left(p_\mathrm{x}(k)\right) 
    & = 0.04537p_\mathrm{x}(k)-0.00576p_\mathrm{x}(k)^2\\
    & \quad{} -0.00143p_\mathrm{x}(k)^3+0.00005p_\mathrm{x}(k)^4\\
    & \quad{} +0.00001p_\mathrm{x}(k)^5,
\end{split}
\end{equation}    
\begin{equation}
\begin{split}
    \mathit{Let:}\quad{} \, p_\mathrm{y}(k) 
    &:= \tilde{a}_\mathrm{y'}(k) - b_\mathrm{ay'},\\
    S_\mathrm{ay'}\left(p_\mathrm{y}(k)\right) &= 0.12723p_\mathrm{y}(k) -0.05823p_\mathrm{y}(k)^2\\
    & \quad{} + 0.00930p_\mathrm{y}(k)^3 -0.00068p_\mathrm{y}(k)^4\\
    & \quad{} + 0.00002p_\mathrm{y}(k)^5.
\end{split}  
\end{equation}

The trajectories of the real values, the measurements without the scale--factor error correction (bias only), and the values after correction with the obtained functions of acceleration are shown in Figs. \ref{fig:blad_nieliniowosciPOZIOM} and \ref{fig:blad_nieliniowosciPION} for $\mathrm{x'}$ and $\mathrm{y'}$ axis, respectively. The $\mathrm{MSE}$ values between the uncorrected measurements and the real trajectories are $0.04817 ~\mathrm{m/s^2}$ and $0.03465 ~\mathrm{m/s^2}$ for $\mathrm{x'}$ and $\mathrm{y'}$ axis, respectively. Whereas, between the corrected measurements and the real trajectories are $0.02720 ~\mathrm{m/s^2}$ and $0.02898 ~\mathrm{m/s^2}$ for $\mathrm{x'}$ and $\mathrm{y'}$ axis, respectively.

In turn, the time constants $T_\mathrm{\dot{\phi}}$ and $T_\mathrm{V}$ of the low--pass filters have been selected by solving an optimisation task providing minimum of $\mathrm{MSE}$ between the reference trajectory of $\phi(\cdot)$ and the calculated angular position $\overline{\phi}(\cdot)$ for each sampling time separately (see Table \ref{tab:small1}). 

\begin{figure}[!ht]
    \centering
    \includegraphics[width=0.99\columnwidth]{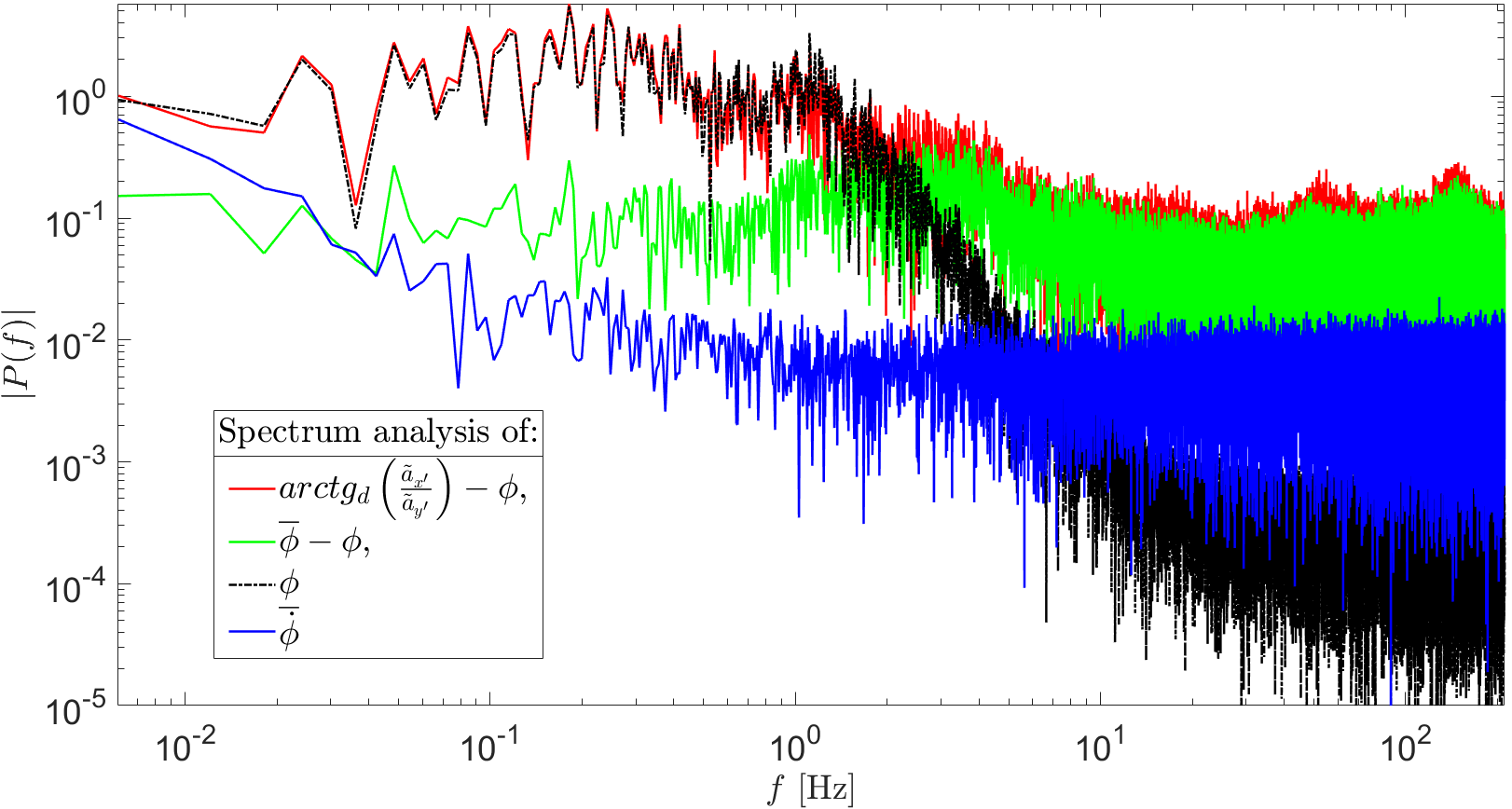}
    \caption{Spectrum of noises}\label{fig:spectrum}
\end{figure}

Next, a noise analysis overlapping $\overline{\phi}$ and $\overline{\dot{\phi}}$ has been performed to check the performance of the correction part. The real angle of tilt from the vertical axis of the two--wheeled balancing robot $\phi$ has been subtracted from the corrected angular position measurement $\overline{\phi}$. Whereas, for the analysis of $\overline{\dot{\phi}}$, the trajectory from Fig. \ref{fig:biasGyro} has been used, because the real angular velocity of the robot is known for this trajectory, and is equal to zero. The obtained noise trajectories have been subjected to fast Fourier transform to obtain the spectrum presented in Fig. \ref{fig:spectrum}. Also, the spectrum of the noise calculated directly from the measurements (without correction) and the real values are shown in Fig. \ref{fig:spectrum}. 
Part of the spectrum analysis of $\phi$ and $\mathrm{arctg}_\mathrm{d}\left(\frac{\tilde{a}_\mathrm{x'}}{\tilde{a}_\mathrm{y'}} \right)-\phi$ overlap, which indicates the influence of $\phi$ on $\mathrm{arctg}_\mathrm{d}\left(\frac{\tilde{a}_\mathrm{x'}}{\tilde{a}_\mathrm{y'}} \right)-\phi$. It can be noticed that the interference caused by changes in the angle of tilt of the robot $\phi$ has been rejected by the correction - there is no overlap in $\phi$ and $\overline{\phi} - \phi$. Thus, Fig.~\ref{fig:spectrum} together with table \ref{tab:small1} testify to the positive influence of the correction part. Moreover, it can be deduced from Fig. \ref{fig:spectrum} that the angular velocity measurement $\overline{\dot{\phi}}$ is much less noisy than the angular position measurement $\overline{\phi}$. The field under the angular velocity noise trajectory equals 1.13, while the field under the trajectory of the real angular velocity $\dot{\phi}$ is equal to 0. Whereas the field under the angular position noise trajectory is 11.42, while the signal to noise ratio is $6.3 ~\mathrm{dB}$. However, basing the filtering system only or mainly on angular velocity may lead to long or no convergence of the angle of tilt estimates $\hat{\phi}$ to the real value in situations where the initial value of angle of tilt is unknown and its current estimate is subject to significant error. The simplest and at the same time the least computationally expensive method of solving this problem the appropriate filter initialisation. During the experiments, the values of initial angular position and velocity have been calculated directly from the measurements as $\hat{\phi}(k=0) = \mathrm{arctg}_\mathrm{d}\left(\frac{\tilde{a}_\mathrm{x'}(k=0)}{\tilde{a}_\mathrm{y'}(k=0)}\right)$, $\hat{\dot{\phi}}(k=0) = \overline{\dot{\phi}}(k=0)$, $\hat{\ddot{\phi}}(k=0)=0$, $\hat{b}_{\dot{\phi}}(k=0)=b_{\dot{\phi}}$. Whereas the parameters of correction parts have been initialised by $\overline{\dot{\phi}}_\mathrm{r,f}(k=0) = \overline{\dot{\phi}}(k=0)$ and $V_\mathrm{t,f}(k=0) = 0$. 

\begin{table*}[!ht]
\small
\centering
 \caption{The obtained values for the low--pass filters}
  \label{tab:small1}
\begin{tabular}{|c|c|c|c|c|}

 \hline
 $\Delta t ~\mathrm{[ms]}$ & $T_\mathrm{\dot{\phi}}$ & $T_\mathrm{V}$ & $\mathrm{MSE}$ for $\hat{\phi}(k) = \mathrm{arctg}_\mathrm{d} 
    \left(
        \frac{\tilde{a}_\mathrm{x'}(k)}{\tilde{a}_\mathrm{y'}(k)} 
    \right)$ & $\mathrm{MSE}$ after correction $\overline{\phi}(k)$\\
 \hline
$2$ & $0.06874$ & $0.04607$ & $150.56951$ & $72.52314$  \\
$5$ & $0.02392$ & $0.02031$ & $515.27065$ & $222.05807$ \\
$10$ & $0.02557$ & $0.02045$ & $377.39749$ & $174.45667$ \\
$20$ & $0.00774$ & $-0.00065$ & $280.23692$ & $110.55792$ \\
\hline
\end{tabular}
\end{table*}

\begin{table*}[!ht]
\small
\centering
\caption{The obtained values for the selected filters}
\label{tab:big1}
\begin{tabular}{|c|c|c|c|c|}
 \hline
 Filter name &$\Delta t ~\mathrm{[ms]}$ & Parameters & $\mathrm{MSE}$ -- `Training' & $\mathrm{MSE}$ -- Verification\\
 \hline
$\alpha$--$\beta$ -- WOB & $2$ & $\alpha = 0.00227$ , $\beta = 1.58242$ & $1.98686$ &  $0.82071$\\
& $5$ & $\alpha = 0.00866$ , $\beta = 1.12381$ & $6.18150$ & $3.07579$\\ 
& $10$ & $\alpha = 0.00103$ , $\beta = 1.67836$ & $1.60046$ & $11.94604$\\
& $20$ & $\alpha = 0.00165$ , $\beta = 1.84408$ & $2.32469$ & $3.34761$\\
\hline
$\alpha$--$\beta$ -- WB & $2$ & $\alpha = 0.00185$ , $\beta = -0.00018$ & $1.93816$ & $0.78603$\\
& $5$ & $\alpha = 0.00858$ , $\beta = -0.00007$ & $6.16623$ & $3.05931$\\
& $10$ & $\alpha = 0.00080$ , $\beta = 0$ & $1.73683$ & $14.17050$\\
& $20$ & $\alpha = 0.00171$ , $\beta = 0$ & $3.07329$ & $4.05407$\\
\hline
$\alpha$--$\beta$--$\theta$--$\gamma$ & $2$ & $\alpha = 0.00204$ , $\beta = -0.00001$ , $\theta = 1.07026$ , $\gamma = -0.00013$ & $0.74852$ & $1.33160$\\
& $5$ & $\alpha = 0.00668$ , $\beta = -0.00005$ , $\theta = 1.05866$ , $\gamma = 0.00007$ & $3.91003$ & $2.12875$\\
& $10$ & $\alpha = 0.00088$ , $\beta = 0$ , $\theta = 1.05141$ , $\gamma = -0.00002$ & $0.61819$ & $12.05841$\\
& $20$ & $\alpha = 0.00391$ , $\beta = -0.00406$ , $\theta = -0.04194$ , $\gamma = 1.87665$ & $2.32142$ & $3.34578$\\
\hline
$\alpha$--$\beta$--$\theta$ -- WA-a & $2$ & $\alpha = 0.00169$ , $\beta = 1.21567$ , $\theta = 0$ & $1.94261$ & $0.86258$\\
& $5$ & $\alpha = 0.00850$ , $\beta = 1.12964$ , $\theta = 0$ & $6.16275$ & $3.07526$\\
& $10$ & $\alpha = 0.00080$ , $\beta = 1.67821$ , $\theta = 0$ & $1.58494$ & $13.92453$\\
& $20$ & $\alpha = 0.00165$ , $\beta = 1.84410$ , $\theta = 0$ & $2.32469$ & $3.34753$\\
\hline
$\alpha$--$\beta$--$\theta$ -- WA-b & $2$ & $\alpha = 0.00315$ , $\beta = 0.28647$ , $\theta = 0.00673$ & $1.87487$ & $1.12021$\\
& $5$ & $\alpha = 0.00911$ , $\beta = 0.32710$ , $\theta = 0.01188$ & $5.43600$ & $2.81706$\\
& $10$ & $\alpha = 0.00104$ , $\beta = 0.69743$ , $\theta = 0.06346$ & $1.33276$ & $11.34403$\\
& $20$ & $\alpha = 0.00168$ , $\beta = 1.01622$ , $\theta = 0.17281$ & $2.06774$ & $3.07125$\\
\hline
Kalman & $2$ & $q_1 = 0.01076$ , $q_2 = 0$ , $r = 0.02792$ & $6.88674$ & $11.08092$\\
& $5$ &  & $9.45858$ & $6.80941$\\
& $10$ & & $9.41660$ & $6.79686$\\
& $20$ & & $7.98366$ & $12.38334$\\
\hline
Kalman* & $2$ & $q_1 = 0.00001$ , $q_2 = 0$ , $r = 2.30640$ & $1.94297$ & $0.79206$\\
& $5$ & $q_1 = 0.00112$ , $q_2 = 0$ , $r = 17.16979$ & $6.17602$ & $3.03979$\\
& $10$ & $q_1 = 0$ , $q_2 = 0$ , $r = 2.25847$ & $1.73928$ & $13.73496$\\
& $20$ & $q_1 = 0.00001$ , $q_2 = 0$ , $r = 2.92997$ & $3.07025$ & $4.10513$\\
\hline
Complementary  & $2$ & $T_\mathrm{c} = 1.06895$ & $2.01177$ &  $0.82619$\\
& $5$ & $T_\mathrm{c} = 0.60307$ & $6.39301$ & $3.38819$\\
& $10$ & $T_\mathrm{c} = 9.74413$ & $1.92216$ & $12.33891$\\
& $20$ & $T_\mathrm{c} = 12.40721$ & $3.31344$ & $4.30363$\\
\hline
\end{tabular}
\end{table*}

\begin{figure*}[!ht]
    \centering
    \includegraphics[width=0.99\textwidth]{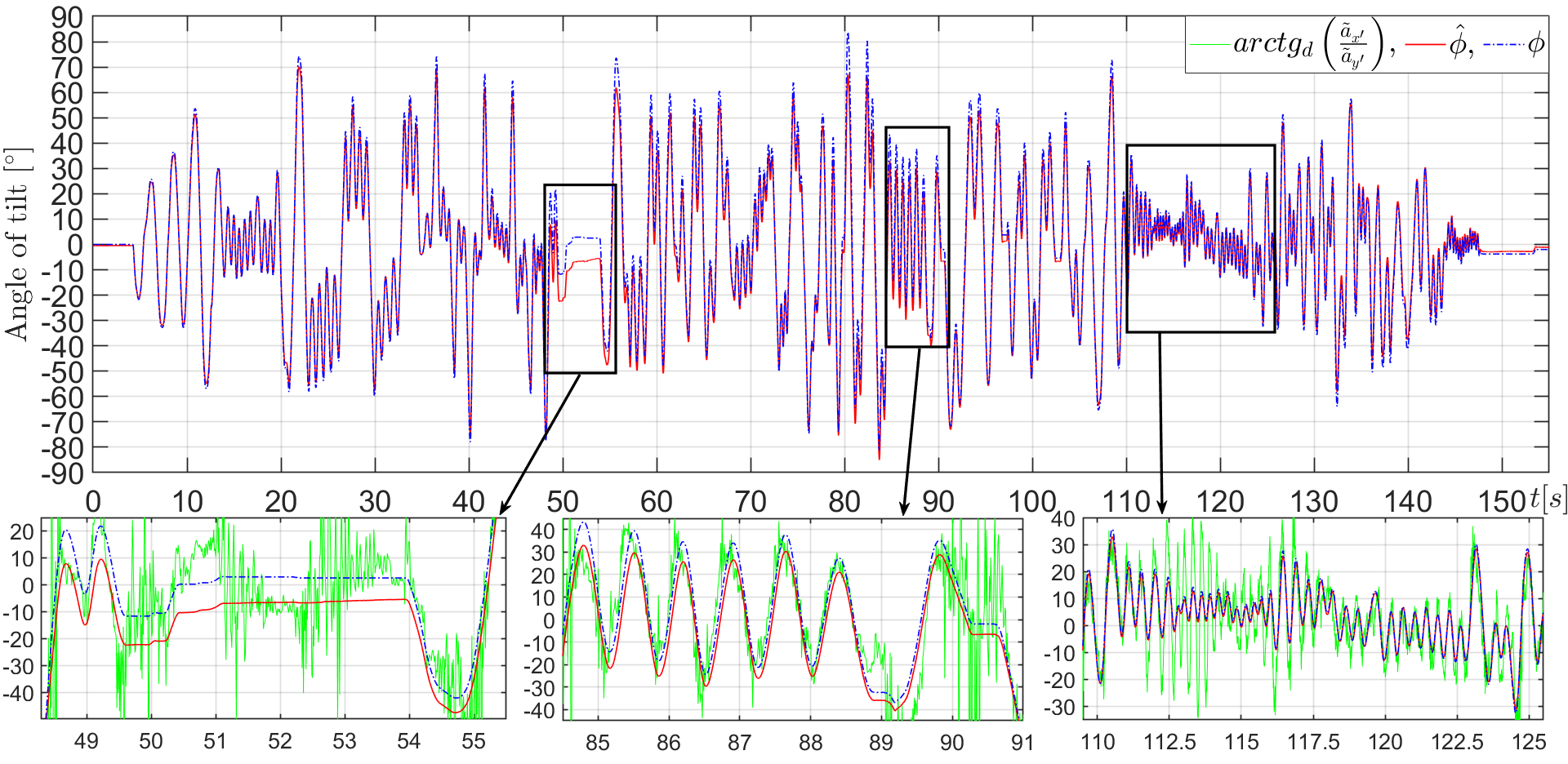}
    \caption{The trajectories of the real and estimated angular position of the two--wheeled balancing robot for $\alpha$--$\beta$ -- WB filter with $\Delta t = 10 ~\mathrm{ms}$}\label{result:fig2}
\end{figure*}

The entire low--cost measurement system shown in Fig. \ref{fig:struktura} has been implemented in research rig presented in Fig. \ref{fig:research_stand}. The results obtained are as follows. Table \ref{tab:small1} shows the obtained value of the low-pass filters constants (equations \eqref{eq:dolnoprzepustowy} and \eqref{eq:filtrPredkosciPost}) and obtained values of $\mathrm{MSE}$ before and after correction part. The values of the parameters of the relevant filters and obtained $\mathrm{MSE}$ values (equation \eqref{eq:MSE}) for the trajectory used in solving the optimisation task (further referred to as `Training') and for the verification trajectory are shown in Table~\ref{tab:big1}. To investigate the real operating conditions and to demonstrate a system performance during dynamic changes and disturbances of the two--wheeled balancing robot movement, the training and verification trajectories have been registered. This took place while the robot moved in both (dynamical) progressive and angular motions. In turn, in order to investigate the performance of the proposed measurement system (with $\alpha$--$\beta$ -- WB filter) the trajectories of the real and estimated angular position of the two--wheeled balancing robot are shown in Fig. \ref{result:fig2}. Moreover, the trajectory of $\mathrm{arctg}_\mathrm{d}\left(\frac{\tilde{a}_\mathrm{x'}}{\tilde{a}_\mathrm{y'}} \right)$ is also presented in Fig.~\ref{result:fig2}. The $\alpha$--$\beta$ -- WB filter has been selected for presentation purposes due to the highest $\mathrm{MSE}$ value in verification -- see Table~\ref{tab:big1}. Thus, the results obtained are satisfactory for the two--wheeled balancing robot stabilisation purposes. However, the performance of both considered Kalman filters is average at best. It is because Kalman filter provides optimal estimates under certain assumptions. Hence, the tests carried out indicate that in the case under consideration these assumptions have not been met. It is particularly important in the body of numerous applications of the `classical' Kalman filter, e.g., in engineering works in the task of angular position estimation from MEMS IMU measurements. On the other hand, considering either of the non-Kalman filters results in comparable estimate precision.%
    \section{Conclusions}\label{sec:conclusions}
In this paper, a low--cost measurement system using filtering of measurements for the two--wheeled balancing robot stabilisation purposes has been investigated. The proposed measurement system includes two layers. First, the physical layer consists of the gyroscopes and accelerometers in MEMS technology and the additional encoder. Second, the software layer containing the correction and filtration mechanisms. The measurements correction is based on the additional encoder, whereas the selected filters, i.e. Kalman, $\alpha$--$\beta$ type, and complementary have been used as the filtration mechanism. The performance of the proposed measurement system has been successfully demonstrated in the experimental setting on the constructed two--wheeled balancing robot. Moreover, the quantitative assessment using a typical measure, i.e. mean square error of selected filters has been provided.%

Hence, in general, extensive knowledge about a low--cost measurement system using filtering of measurements for the two--wheeled balancing robot stabilisation purposes has been aggregated in this paper. It can be found interesting and useful for the relevant community, both in research and engineering applications.%

The future research may be the conditioning of measurement signals in such a way as to meet the assumptions of Kalman filter or the use of more complex and computationally expensive filters, e.g., extended Kalman filter or adaptive-type Kalman filters.
\section*{Acknowledgement}
The research was done in accordance with funding from Polish MEiN under Young Researcher Support Program. The authors wish to express their thanks for the support.
    
    \bibliographystyle{model3-num-names}%
    \bibliography{sections/references}%
\bio{img/KL_photo}%
Krzysztof Laddach received the M.Sc. degree (Hons.) in control engineering from the Faculty of Electrical and Control Engineering, the Gda{\'n}sk University of Technology in 2019. His master’s dissertation was focused on algorithms of artificial neural network structure search for processes dynamics modelling purposes. Since October 2019 he has been a Ph.D. student in the Doctoral School at the Gda{\'n}sk University of Technology. His research interests involve mathematical modelling, especially in connection with the use of computational intelligence (artificial intelligence), optimal selection of the architecture of artificial neural networks, and the use of artificial neural networks in estimation.%
\endbio%
\bio{img/RL_photo}%
Rafa{\l} {\L}angowski received the M.Sc. and the Ph.D. degrees (Hons.) in control engineering from the Faculty of Electrical and Control Engineering at the Gda{\'n}sk University of Technology in 2003 and 2015, respectively. From 2007 to 2014, he held the specialist as well as manager positions at ENERGA, one of the biggest energy enterprises in Poland. Since February 2014, he has been an owner of VIDEN a business in energy and control areas. He provides theoretical and practical experience, especially in front and back office at energy company and operation of the energy market in Poland. From 2016 to 2017, he was a Senior Lecturer with the Department of Control Systems Engineering at the Gda{\'n}sk University of Technology. He is currently an Assistant Professor with the Department of Electrical Engineering, Control Systems and Informatics. His research interests involve mathematical modelling and identification, estimation methods, especially state observers, and monitoring of large scale complex systems.%
\endbio%
\bio{img/TZ_photo}%
Tomasz Zubowicz received his M.Sc. Eng. in control engineering from the Faculty of Electrical and Control Engineering at the Gda{\'n}sk University of Technology in 2008. He received his Ph.D. Eng. (Hons.) in the field of control engineering from the same faculty in 2019. In 2012, he became a permanent staff member at the Department of Control Systems Engineering at Gda{\'n}sk University of Technology and a member of the IFAC T.C. 5.4 Large Scale Complex Systems. He currently holds position of an Assistant Professor with the Department of Electrical Engineering, Control Systems and Informatics at Gda{\'n}sk University of Technology and serves as a Head Specialist for Space Situation Awareness Team at Polish Space Agency. His current research interests concern monitoring, control and security of critical infrastructure systems.%
\endbio%

\end{document}